\documentclass[10pt,technote,compsoc]{IEEEtran}

\ifCLASSOPTIONcompsoc
  \usepackage[nocompress]{cite}
\else
  \usepackage{cite}
\fi
\ifCLASSINFOpdf
\else
\fi

\usepackage{graphicx}
\usepackage{amsmath}
\usepackage{amssymb}
\usepackage{booktabs}
\usepackage{color}
\usepackage{mathtools}
\usepackage{amsfonts}
\usepackage{bbm}
\usepackage{bm}
\usepackage{makecell}

\begin{document}
%
\title{One-Shot Action Recognition via Multi-Scale Spatial-Temporal Skeleton Matching}
%
%
%
%

\author{Siyuan Yang,
        Jun Liu,
        Shijian Lu,
        Er Meng Hwa,~\IEEEmembership{Life~Fellow,~IEEE},
        and~Alex C. Kot,~\IEEEmembership{Life~Fellow,~IEEE}
\IEEEcompsocitemizethanks{
\IEEEcompsocthanksitem Siyuan Yang is with the Rapid-Rich Object Search Lab, Interdisciplinary Graduate Programme, Nanyang Technological University, Singapore.\protect\\
E-mail: SIYUAN005@e.ntu.edu.sg
\IEEEcompsocthanksitem Jun Liu is with the Information Systems Technology and Design Pillar, Singapore University of Technology and Design, Singapore.\protect\\
E-mail: jun\_liu@sutd.edu.sg 
\IEEEcompsocthanksitem Shijian Lu is with the School of Computer Science \& Engineering, Nanyang Technological University, Singapore.\protect\\
E-mail: Shijian.Lu@ntu.edu.sg 
\IEEEcompsocthanksitem Er Meng Hwa and Alex C. Kot are with the School of Electrical and Electronic Engineering, Nanyang Technological University, Singapore.\protect\\
E-mail: \{emher, eackot\}@ntu.edu.sg
}
\thanks{Corresponding author: Jun Liu}
}

%
%

\markboth{TPAMI - Short Paper Submission}%
{Shell \MakeLowercase{\textit{et al.}}: Bare Demo of IEEEtran.cls for Computer Society Journals}
%



\IEEEtitleabstractindextext{%
\begin{abstract}
One-shot skeleton action recognition, which aims to learn a skeleton action recognition model with a single training sample, has attracted increasing interest due to the challenge of collecting and annotating large-scale skeleton action data. However, most existing studies match skeleton sequences by comparing their feature vectors directly which neglects spatial structures and temporal orders of skeleton data. 
This paper presents a novel one-shot skeleton action recognition technique that handles skeleton action recognition via multi-scale spatial-temporal feature matching.
We represent skeleton data at multiple spatial and temporal scales and achieve optimal feature matching from two perspectives. The first is multi-scale matching which captures the scale-wise semantic relevance of skeleton data at multiple spatial and temporal scales simultaneously. The second is cross-scale matching which handles different motion magnitudes and speeds by capturing sample-wise relevance across multiple scales. Extensive experiments over three large-scale datasets (NTU RGB+D, NTU RGB+D 120, and PKU-MMD) show that our method achieves superior one-shot skeleton action recognition, and outperforms SOTA consistently by large margins.
\end{abstract}

}

\maketitle

\IEEEdisplaynontitleabstractindextext

%
\IEEEpeerreviewmaketitle


%
%
%
%
\section{Introduction}
\label{intro}
Human action recognition is a fast-developing research area due to its wide applications in human-computer interaction, video surveillance, game control, etc. 
In recent years, human action recognition with skeleton data has attracted increasing attention as skeleton data encodes high-level representations of human actions and is generally lightweight and robust to variations in appearances, surrounding distractions, viewpoint changes, etc.
As of today, most existing studies expect large-scale labeled training data for learning effective human action representations. While facing skeleton data of a new category, they require to collect hundreds of action samples of the new category for adapting or fine-tuning some existing models.
How to achieve single-shot recognition for new action categories becomes critically important for circumventing the tedious and laborious data collection and labeling procedure.

\begin{figure}[t]
\begin{center}
\includegraphics[trim=0.2cm 0cm 0.1cm 0.2cm,clip, width=0.45\textwidth]{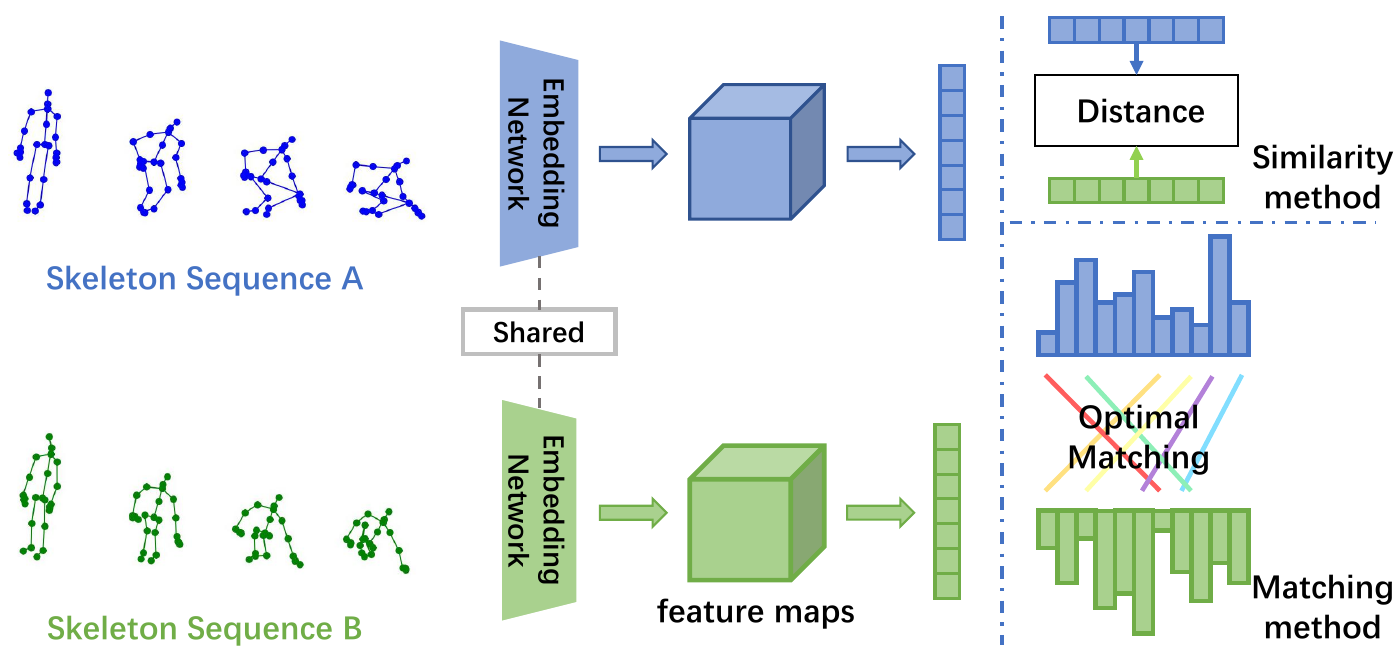}
\end{center}
\vspace{-0.3cm}
  \caption{
Skeleton action recognition based on feature similarity or feature matching: Feature similarity computes the distance between feature vectors which discards the very useful spatial skeleton structures and temporal information. The proposed feature matching compares two skeleton sequences by computing a matching flow between their feature distributions which can capture useful spatial and temporal information effectively. {The colored line emphasizes channels paired based on their high matching scores.}
 }
\vspace{-0.6cm}
\label{fig:framework compare}
\end{figure}

One-shot skeleton action recognition is a very challenging task. Beyond the data constraint with one-shot for unseen new classes, the major challenge comes from the very rich variations in human actions. Take the action ``put on glasses'' as an example. Different persons could perform it by using their left hand, right hand, or both hands. The same person could also perform it at different paces with different motion dynamics.
Different approaches have been explored to address this challenging task, and most existing works~\cite{ntu120,memmesheimer2020skeleton,memmesheimer2020signal, sabater2021one} represent the anchor and target samples with certain pooled feature vectors and compute the sample distance based on the similarity of their pooled feature vectors. However, the adoption of such global feature similarity discards the very useful spatial structures and temporal order of the skeleton sequences. In addition, most existing works learn skeleton action representations at a single scale of the original body joints which tends to lose useful action features under the one-shot scenario. 
{Drawing from ~\cite{cai2019exploiting,yang2020collaborative,10288273,chen2021multi,liu2020disentangling}, it is evident that human actions are multi-scales in both spatial and temporal spaces.
For instance, multiple joints on arms and legs collaborate in walking, and consecutive frames of human actions contain strong temporal correlations.
Skeleton action representations should therefore capture the rich semantic correlations at different spatial and temporal scales. 
In contrast to~\cite{cai2019exploiting,yang2020collaborative,10288273,chen2021multi,liu2020disentangling}, which employ multi-scale information for recognition and prediction, we leverage the extracted multi-scale features for match and design the innovative matching strategies specifically for one-shot skeleton action recognition challenge.
}

\begin{figure}[t]
\begin{center}
\includegraphics[trim=0cm 0.1cm 0.1cm 0cm,clip,width=0.33\textwidth]{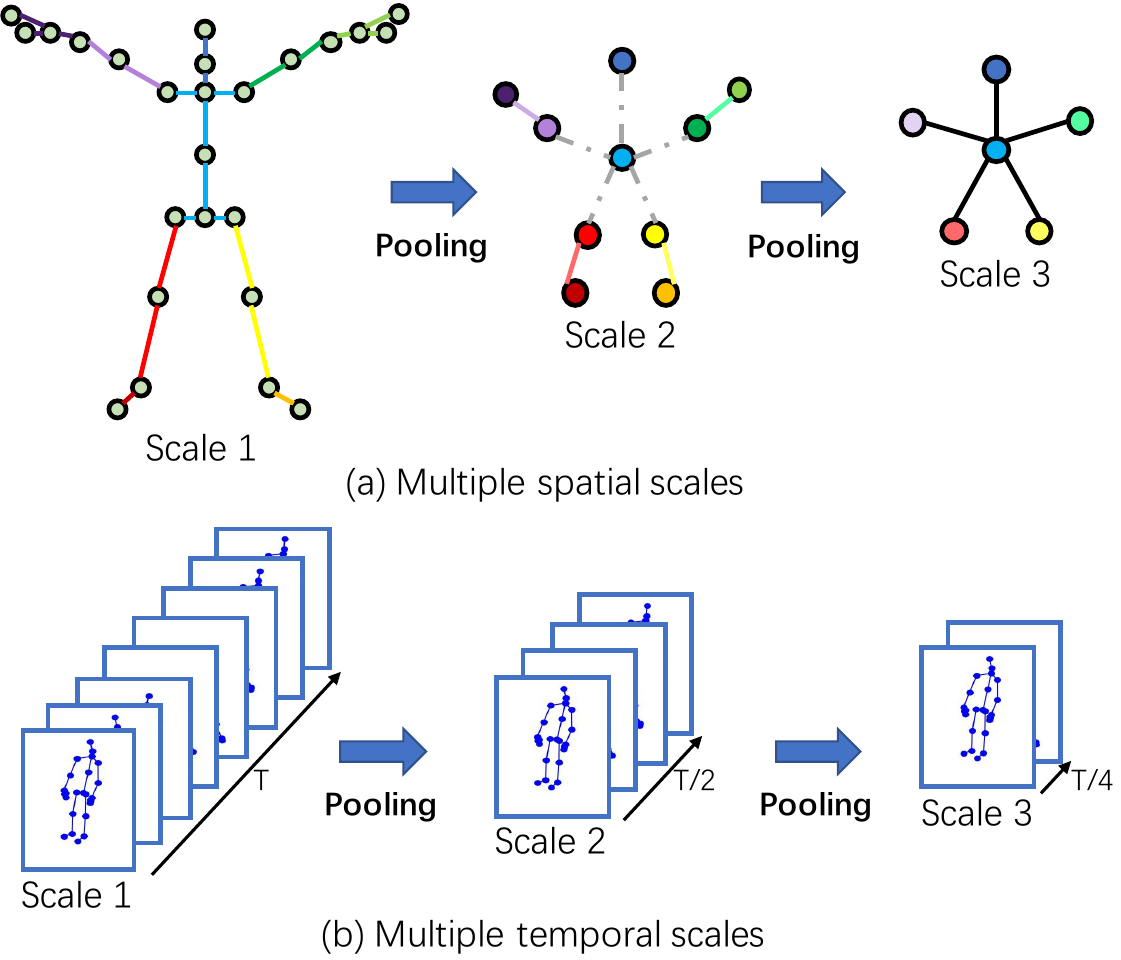}
\vspace{-0.5cm}
\end{center}
  \caption{
   The proposed multi-scale skeleton modeling at spatial dimension in (a) and temporal dimension in (b): Given the original spatial scale at \textit{Scale 1}
   , we first divide the skeleton nodes into multiple groups with similar semantic skeleton structures and then perform average pooling to each group to generate skeleton graphs of coarser scales
   (i.e., \textit{Scale 2} and \textit{Scale 3}.)
   The nodes whose links are of the same color belong to the same group with similar semantics. Along the temporal dimension, we perform average pooling over features of adjacent frames to obtain temporal features of coarser scales at \textit{Scale 2} and \textit{Scale 3}.
   (Details of spatial-pooling are available in the appendix.) 
  }
\vspace{-0.6cm}
\label{fig:pooling}
\end{figure}

We propose to capture spatial-temporal features by leveraging spatial structures and temporal orders of skeleton sequences as illustrated in Fig.~\ref{fig:framework compare}. Inspired by the theory of optimal transport~\cite{cuturi2013sinkhorn,villani2009optimal}, we measure the semantic relevance of two skeleton sequences by computing an optimal matching flow between their feature maps. 
{Specifically, we adopt Earth Mover's Distance (EMD)~\cite{rubner2000earth} as the optimal matching metric for acquiring the optimal matching flow. 
EMD is the metric for computing the distance between two representations, enabling us to determine the similarity between the feature representations of two skeleton samples.
In our one-shot skeleton action recognition scenario, given the distances between all skeleton joint pairs, EMD maximizes the impact of relevant joints and minimizes the effect of irrelevant joints between two skeleton sequences.} 
In addition, we model skeleton sequences at multiple spatial scales (joint-scale, part-scale, and limb-scale) and temporal scales as illustrated in Fig.~\ref{fig:pooling}, and perform multi-scale matching to capture scale-wise skeleton semantic relevance by using EMD. Further, human action could be performed with different motion magnitudes and motion paces, e.g., `hand waving' may be performed by hand (joint-level), forearm (part-level), or the whole arm (limb-level) at different paces. We thus design cross-scale matching that learns semantic relevance by measuring feature consistency across spatial and temporal scales.

The contributions of this work are threefold. 
\textit{First}, we formulate one-shot skeleton action recognition as an optimal matching problem and design an effective network framework for one-shot skeleton action recognition. 
\textit{Second,}, we propose a multi-scale matching strategy that can capture scale-wise skeleton semantic relevance at multiple spatial and temporal scales. On top of that, we design a novel cross-scale matching scheme that can model the within-class variation of human actions in motion magnitudes and motion paces. To the best of our knowledge, this is the first work that exploits multi-scale representations and cross-scale matching to capture multi-scale skeleton semantic relevance and maintain consistency across motion scales in one-shot skeleton action recognition. \textit{Third,} extensive experiments on three public datasets (NTU RGB+D, NTU RGB+D 120, and PKU-MMD) show that our method outperforms the state-of-the-art consistently by large margins.

\vspace{-0.1cm}
\section{Related Works}
{\textbf{Skeleton-based Action Recognition} 
has attracted increasing attention in recent years.
Traditional methods design hand-craft features to represent skeleton sequences \cite{hussein2013human,vemulapalli2014human,wang2013learning}.
In recent years, deep learning has been widely explored 
by leveraging the advances in Recurrent Neural Networks \cite{du2015hierarchical,liu2017skeleton,liu2016spatio}, Convolutional Neural Networks \cite{du2015skeleton,ke2017new,li2017skeleton,LiChao}, Graph Convolution Networks (GCNs) \cite{cheng2020skeleton,liu2020disentangling,Shi_2019_CVPR_twostream,yan2018spatial}, Hypergraph Neural Networks~\cite{Wang_2023_CVPR, zhu2022selective,hao2021hypergraph}, and Transformer-based methods~\cite{Wang_2023_CVPR, ahn2023star, bertasius2021space}.
Given the inherent topological graph structure of the human skeleton,
GCNs have attracted increasing attention in skeleton-based action recognition. 
Notably,
Yan \textit{et al}. \cite{yan2018spatial} proposed a spatial-temporal graph convolutional network to learn spatial-temporal patterns from skeleton data.
Shi \textit{et al}. \cite{Shi_2019_CVPR_twostream} designed an adaptive graph convolutional network, which utilized self-attention with the spatial-temporal GCN.
In a more recent development, Wang \textit{et al}. \cite{wang2023neural} employed parameterized Koopman pooling, replacing average pooling in supervised skeleton action recognition.}

Though the prior studies achieved very impressive performance, most of them are supervised and require large-scale training data which is often laborious to collect.
We focus on one-shot skeleton action recognition, aiming to address the scenarios where only a few labeled skeleton samples are available for an unseen new class.

\noindent\textbf{Few-Shot Learning.}
Motivated by the human capability in learning new concepts from just a few samples, few-shot learning,
which aims at recognizing unseen concepts with only a few labeled training samples, has received increasing attention and witnessed significant advances in recent years. 
It has also been widely explored in the computer vision research community. For example,
Snell \textit{et al.}~\cite{snell2017prototypical} presented the Prototypical Networks that compute distances between a datapoint and class-wise prototypes.
Ye \textit{et al.} \cite{ye2020fewshot} defined set-to-set transformations to learn a task-specific feature embedding for few-shot learning.
Simon \textit{et al.} \cite{simon2020adaptive} presented the DSN that employs a few-shot learning model via affine subspaces.
Xu \textit{et al.} \cite{xu2021learning} proposed to learn a novel dynamic meta-filter for few-shot learning.

\noindent{\textbf{One-shot Skeleton Action Recognition}
has attracted increasing interest in
recent years.
Leveraging the NTU RGB+D 120 dataset, Liu \textit{et al.} \cite{ntu120} first presented
an Action-Part Semantic-Relevance aware (APSR) approach for one-shot skeleton action recognition.
Sabater \textit{et al.} \cite{sabater2021one} presented a one-shot action recognition approach based on a Temporal Convolutional Network (TCN).
Memmesheimer \textit{et al.} \cite{memmesheimer2020signal} proposed to formulate the one-shot skeleton action learning problem as a deep metric learning problem.
Additionally, Memmesheimer \textit{et al.} \cite{memmesheimer2020skeleton} presented an image-based skeleton representation, which performs well in the deep metric learning manner.
Ma \textit{et al.}~\cite{ma2022learning} proved that maximally preserving disentangled joint-level spatial features are beneficial to increase representation diversity and recognizability for few-shot classes in one-shot skeleton action recognition.
Wang \textit{et al.}~\cite{wang2022temporal} proposed JEANIE, which performs the joint alignment of temporal blocks and simulated viewpoint indexes of skeletons between support-query sequences to select the smoothest path without abrupt jumps in matching temporal locations and view indexes.
In a more recent study, Wang \textit{et al.}~\cite{wang2022uncertainty} introduced the uncertainty-DTW, which take into account the uncertainty of in frame-wise (or block-wise) features by selecting the path which maximizes the Maximum Likelihood Estimation (MLE).
}

Most existing methods take the skeleton representation as a whole and measure the skeleton similarity globally which often misses useful structure and temporal information. We propose to treat one-shot skeleton action recognition as an optimal matching problem and design multi-scale matching and cross-scale matching which capture the scale-wise semantic relevance and maintain the spatial and temporal consistency across different scales, respectively

\section{Method}
\begin{figure}[t]
\begin{center}
\includegraphics[trim=0.3cm 0.3cm 0.1cm 0.1cm,clip,width=0.43\textwidth]{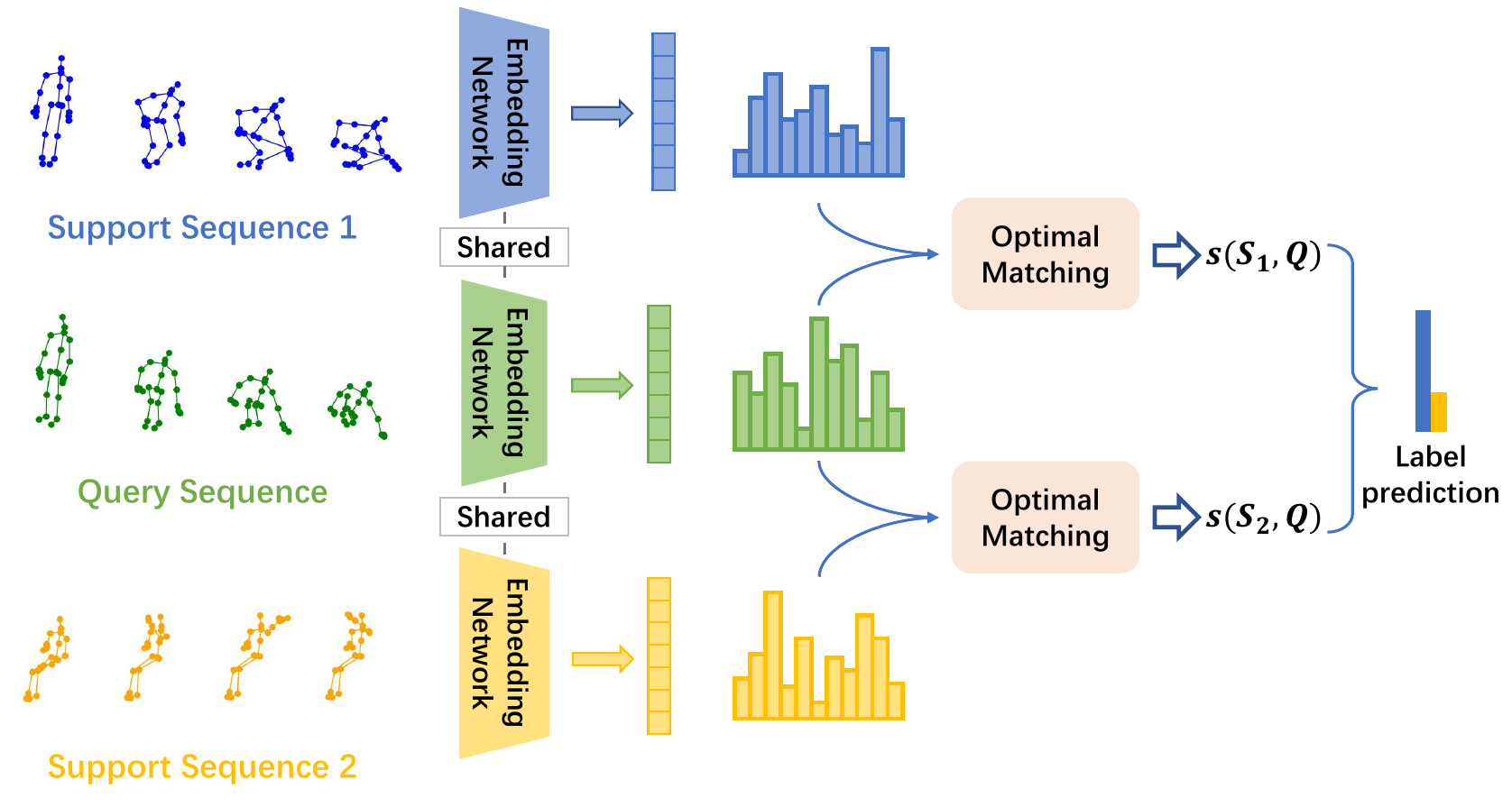}
 \vspace{-0.3cm}
\end{center}
\caption{
The pipeline of the proposed method: Given a \textit{Query Sequence} and two support skeleton sequences \textit{Support Sequence 1} and \textit{Support Sequence 2}, skeleton representations are first extracted with a weight-sharing embedding network which are further aligned progressively with the proposed \textit{Optimal Matching}. The semantic relevance score (denoted by $s(\cdot, \cdot)$) between the query and the two support instances can then be computed for action prediction. The pipeline is illustrated with a 2-way 1-shot task. Different \textit{Optimal Matching} strategies are provided in Fig.~\ref{fig:cross optimal match}.
  }
\vspace{-0.3cm}  
\label{fig:framework}
\end{figure}

We aim to train a model that can recognize human skeleton data of novel classes with only a single labeled sample. It is a very challenging task due to the very rich intra-class spatial-temporal variations in human skeleton action. We address this challenge by proposing a one-shot skeleton action recognition framework as illustrated in Fig.~\ref{fig:framework}. Specifically, we design a novel optimal matching technique to capture the useful spatial structure and temporal order information which is largely neglected in most existing one-shot skeleton action recognition studies.
In the following, we first present the problem formulation of the one-shot skeleton action recognition task. We then introduce the embedding network and elaborate on how to construct multi-spatial and multi-temporal scale skeletons. Finally, we describe the proposed optimal matching technique in detail.
{The important notations and definitions are summarized in Tab.~\ref{tab: notations} in the appendix.
}

\vspace{-0.3cm}  
\subsection{Problem Formulation}
\label{ProblemFormulation}
Inspired by prior studies in few-shot learning \cite{simon2020adaptive,snell2017prototypical, xu2021learning,ye2020fewshot,Zhang_2020_CVPR} and few-shot video action recognition  \cite{bishay2019tarn,cao2020few,ECCV2020fewshot,Wang_2022_CVPR,Thatipelli_2022_CVPR,zhu2018compound}, we formulate the one-shot skeleton action recognition task as a meta-learning problem~\cite{vinyals2016matching} that consists of a meta-training phase and a meta-testing phase. In a \textit{n}-way and \textit{1}-shot problem, each episode consists of a support set \textit{S} and a query set \textit{Q}, where \textit{S} contains \textit{1} labeled sample for each of \textit{n} unseen classes and \textit{Q} is employed to evaluate the generalization performance. The algorithm aims to determine which support classes each query sample belongs to. Specifically, multiple \textit{n}-way and \textit{1}-shot tasks are randomly sampled from the \textit{meta-training set $D_{train}$} (with seen classes), and employed to train a model in an episodic manner. In \textit{meta-testing phase}, \textit{n}-way and \textit{1}-shot tasks are sampled from the \textit{meta-testing set $D_{test}$} (with unseen classes) for evaluations.

\vspace{-0.3cm}  
\subsection{Skeleton Feature Embedding}
\label{Pretraining}

{Following prior studies on few-shot learning \cite{simon2020adaptive,snell2017prototypical,xu2021learning,ye2020fewshot,Zhang_2020_CVPR} and few-shot video action recognition \cite{bishay2019tarn,cao2020few,ECCV2020fewshot,Wang_2022_CVPR,Thatipelli_2022_CVPR,zhu2018compound}, we first pre-train an embedding network on the whole \textit{meta-training set $D_{train}$} using the cross-entropy loss for standard classification before proceeding to episodic training.}
We adopt the GCN-based model~\cite{Shi_2019_CVPR_twostream} as the baseline network which has an adaptive spatial-temporal graph for extracting the relation among body joints. 
The GCN-based model just processes the features of the original scale. However, such single-scale modeling often misses meaningful skeleton information especially when only a single labeled sample is available as described in Sec. \ref{intro}. Inspired by  \cite{cai2019exploiting, newell2016stacked,yang2020collaborative,chen2021multi,liu2020disentangling} which handle multi-scale features, we represent the human skeleton data at multi-spatial and multi-temporal scales.

\noindent\textbf{Multi-Spatial Scale Skeleton:}
We model skeleton actions at multiple spatial scales. Specifically, we adopt 3 spatial scales including the body-joint scale ($s_1$), the part-level scale ($s_2$), and the limb-level scale ($s_3$) as illustrated in Fig.~\ref{fig:pooling} (a). We first build GCN blocks on the first scale to capture joint-wise feature representations and then perform the average pooling ({the details of spatial pooing can be found in Appendix D}). For skeleton-based representations, the pooling requires meaningful neighborhoods and we simply put joints of the same spatial scale into one group.

\noindent\textbf{Multi-Temporal Scale Skeleton:}
{
Recognizing that consecutive frames capture continuous motions and poses reflecting analogous abstract states, we represent skeleton data across multiple temporal scales. 
After processing several GCN blocks at the original temporal scale, we incorporate two average pooling layers along the temporal dimension to perform temporal pooling as illustrated in Fig.~\ref{fig:pooling} (b) and Fig.~\ref{fig:multi-temporal scale}. 
Specifically, by applying average pooling to the features of consecutive frames from the original scale, we consolidate them into a unified feature to represent a `new frame' in coarser scales, such as scale 2 or scale 3
As we apply average pooling with strides of 2 and 4, the features of two consecutive frames are pooled to produce the scale 2 features, while those of four consecutive frames are pooled to yield the scale 3 features.
}

We implement 3 spatial scales and 3 temporal scales for illustration, where each skeleton structure captures unique perspectives of skeleton representations. 
{
To extract the multi-scale skeleton representation,
each stream within the multi-spatial and multi-temporal scale skeleton structures is individually optimized through the cross-entropy loss\footnote{Detailed network structures can be found in the appendix.\label{footnote}}.
}

\vspace{-0.3cm}  
\subsection{Optimal Matching Strategy}
\label{Matching}

Unlike~\cite{ntu120,memmesheimer2020skeleton,memmesheimer2020signal,sabater2021one} that compute distances over sequence-level embeddings, we capture discriminative local information of each body joint and design a skeleton optimal matching scheme to compute semantic relevance based on optimal transport theory as discussed in Sec. \ref{intro}. Specifically, we adopt the Earth Mover’s Distance (EMD) \cite{rubner2000earth} as the optimal transport matching metric, which searches for the minimal cost transport plan between two joints' feature distributions by maximizing the impact caused by relevant joints and minimizing the effect between irrelevant joints.

\begin{figure}[t]
\begin{center}
\includegraphics[trim=0cm 0.2cm 0cm 0cm,clip,width=0.45\textwidth]{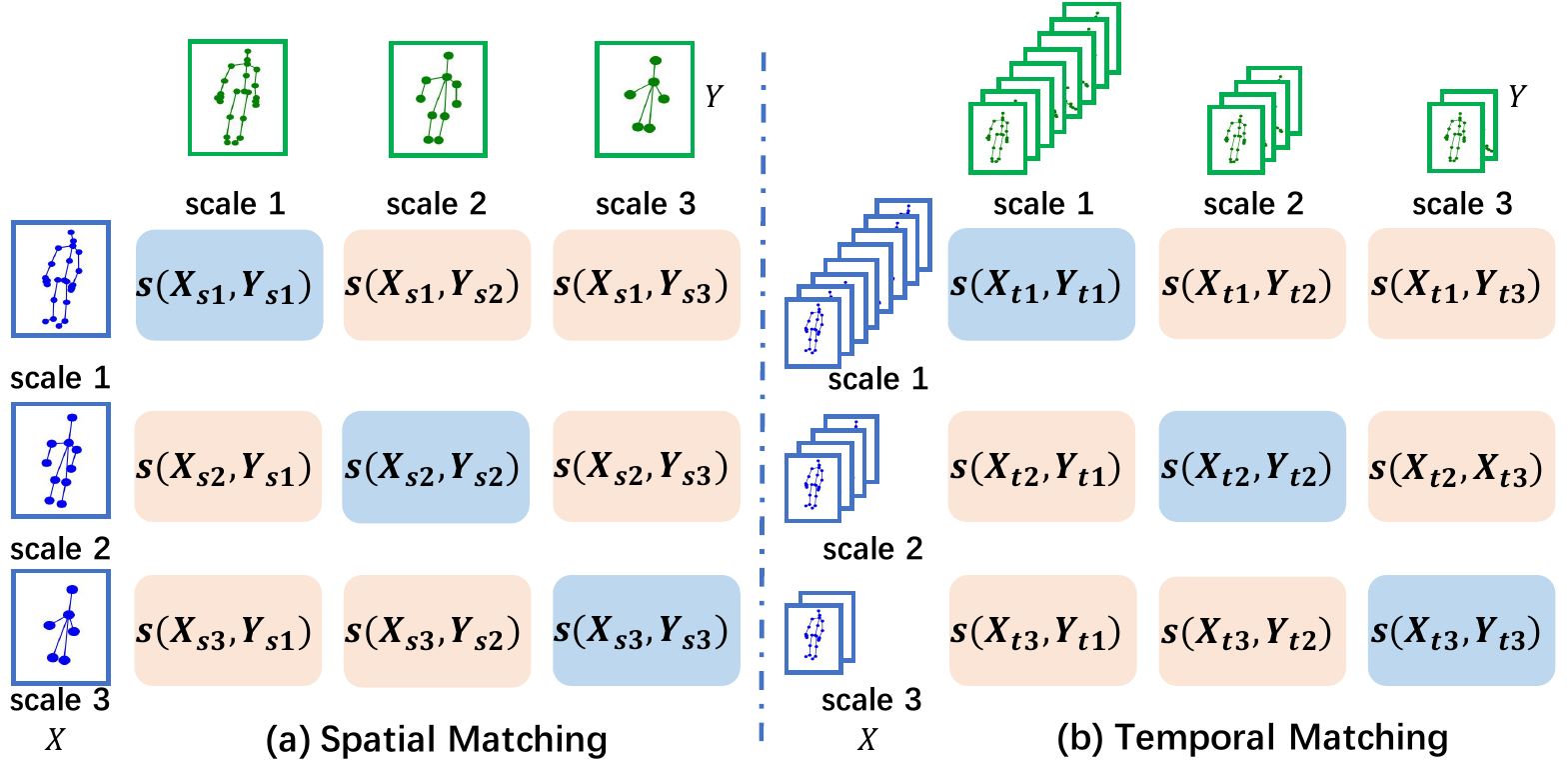}
\vspace{-0.4cm}
\end{center}
  \caption{
   Illustration of optimal matching in Spatial Matching in (a) and Temporal Matching in (b): In each of the two sub-figures, the three matching strategies along the diagonal (in blue color) illustrate the proposed multi-scale matching, and the rest 6 of the diagonal (in orange color) show the proposed cross-scale matching. Here, $s(\cdot, \cdot)$ represents the semantic relevance score between two skeleton features. \textit{X} and \textit{Y} stand for two skeleton sequences. 
  }
\vspace{-0.5cm}  
\label{fig:cross optimal match}
\end{figure}

Here, we first use the single-scale model as an example to show how we formulate the one-shot skeleton action recognition as the optimal matching problem by using EMD. 
The skeleton representation embedded by the single-scale model can be represented as $\textbf{X} \in \mathbbm{R}^{C \times N \times T}$, where $C$ is the number of output channels, $N$ denotes the number of skeleton joints, and $T$ denotes the number of frames. 
For two feature maps $\textbf{X}, \textbf{Y} \in \mathbbm{R}^{C \times N \times T}$, we first flatten them into two sets of joint's local representations $\mathcal{X} = \left\{x_{i}|i = 1, 2, ...NT \right\}$ and $\mathcal{Y}=\left\{y_{j} \mid j=1,2, \ldots NT\right\}$, where $x_{i}$ and $y_{j}$ ($x_{i}, y_{j} \in \mathbbm{R}^C$) denote the joint's local representation at the corresponding spatial and temporal positions. 
Then we define the EMD between two sets of local representations as the minimum “transport cost” from $\mathcal{X}$ (suppliers) to $\mathcal{Y}$ (demanders). 
Suppose for each supplier $x_{i}$, it has $r_{i}$ units to transport, and for each demander $y_{j}$, it requires $c_{j}$ units. 
The overall optimal transport matching problem can be formulated as:
\begin{equation}
\small
    OT(r, c) \:= \left\{\pi \in \mathbbm{R}^{NT \times NT} | \pi  \mathbbm{1} = \mathbf{r}, \pi^{\top} \mathbbm{1} = \mathbf{c} \right\}, \label{emd org}
\end{equation}
\noindent where $\pi$ is the optimal matching flow between these two distributions, which can also be viewed as the optimal matching plan of two skeleton sequences.
$r_i$ and $c_j$ are called the weights of nodes, which control the total matching flows generated by each node, and $\mathbf{r}$ and $\mathbf{c}$ are vectorized representations of $\left \{ r_{i}\right\}$ and $\left \{c_{j}\right\}$.
EMD seeks an optimal matching flow $\pi$ between ``suppliers'' $\mathcal{X}$ and ``demanders'' $\mathcal{Y}$, such that the overall matching cost can be minimized.

Additionally, the transporting cost per unit is defined by computing the pairwise distance between supplier node $\mathbf{x}_{i}$ and demander node $\mathbf{y}_{j}$ from two skeleton features:
\begin{equation}
\small
    d_{i j}=1-\frac{\mathbf{x}_{i}^{\top} \mathbf{y}_{j}}{\left\|\mathbf{x}_{i}\right\|\left\|\mathbf{y}_{j}\right\|}, 
\end{equation}
\noindent where nodes with similar local representations tend to generate small transporting costs between each other. 
Then we can define the EMD as the optimal transportation problem, which is represented as:
\begin{equation}
\small
    D_{emd}(\textbf{X}, \textbf{Y}) =\mathop{min}\limits_{\pi \in OT(r, c)} \sum_{i=1}^{NT}\sum_{j=1}^{NT} d_{ij} \pi_{ij}.
\end{equation}

The weight of each node (\textit{e.g.}, $r_i$ and $c_j$) plays an important role in optimal matching problems. Intuitively, the node with a larger weight is more important while matching two sets.
Therefore, in order to assign the more important node a higher weight, we follow \cite{Zhang_2020_CVPR} to generate the weight $r_{i}$ by a cross-reference mechanism that uses the dot product between a joint representation and the average joint representation in the other skeleton features:
\begin{equation}
\small
    r_{i} = max \left\{x_{i}^{\top} \cdot \frac{\sum_{j=1}^{NT}y_j}{NT}, 0  \right\},
\end{equation}
where $x_{i}$ and $y_j$ denote the feature vectors from two skeleton feature maps, and function $max(\cdot)$ ensures the weight is always non-negative.
Above we take $r_i$ as an example, and $c_j$ can be calculated in the same manner.
Once acquiring the optimal matching flow $\pi$, we can compute the semantic relevance score $s$ between two skeleton representations as:
\begin{equation}
\small
    s(\textbf{X}, \textbf{Y}) = \sum_{i=1}^{NT}\sum_{j=1}^{NT}(1-d_{ij})\pi_{ij}.
    \label{single-scale}
\end{equation}
\noindent These semantic relevance scores allow studying the composition of the overall relevance, enabling us to assign high relevance to semantically similar joints no matter whether they are in the same spatial order or temporal frame. 
We can thus tackle the problem that the semantic relevance of two skeleton sequences can occur at different temporal positions or different spatial joints.

\noindent\textbf{Multi-Scale Matching.}
As mentioned in Sec. \ref{Pretraining}, 
human skeleton data can be represented in multi-spatial scales and multi-temporal scales, and each scale's representation contains unique semantic information.
Thus, we propose to capture the pair-wise skeleton semantic relevance at multiple scales, including multi-spatial scale matching and multi-temporal scale matching, to acquire optimal matching flow from multiple spatial and multiple temporal scales.

For the multi-spatial scale scenario, there are three pairs of feature embeddings, which can be represented as $\textbf{X}_{s1} \in \mathbbm{R}^{C \times N \times T}$, $\textbf{X}_{s2} \in \mathbbm{R}^{C \times N_2 \times T}$, and $\textbf{X}_{s3} \in \mathbbm{R}^{C \times N_3 \times T}$, respectively. 
$N_2$ denotes the number of nodes for the second-scale spatial graph, and $N_3$ stands for the number of third-scale spatial graph nodes.
The semantic relevance score between two skeleton sequences thus becomes:
\begin{equation}
\small
    s_{ms}(\textbf{X}, \textbf{Y}) = s(\textbf{X}_{s1}, \textbf{Y}_{s1}) + s(\textbf{X}_{s2}, \textbf{Y}_{s2}) + s(\textbf{X}_{s3}, \textbf{Y}_{s3}).\label{multi-s emd}
\end{equation}
\noindent  
This enables us to seek the optimal matching flow using EMD, and measure the semantic relevance at multiple spatial scales as shown in Fig.~\ref{fig:cross optimal match} (a) (highlighted in blue color).

For the multi-temporal scale, there are also three pairs of feature embeddings, which can be represented as $\textbf{X}_{t1} \in \mathbbm{R}^{C \times N \times T}$, $\textbf{X}_{t2} \in \mathbbm{R}^{C \times N \times T/2}$, and $\textbf{X}_{t3} \in \mathbbm{R}^{C \times N \times T/4}$, respectively. 
The semantic relevance score between two skeleton sequences becomes:
\begin{equation}
\small
    s_{mt}(\textbf{X}, \textbf{Y}) = s(\textbf{X}_{t1}, \textbf{Y}_{t1}) + s(\textbf{X}_{t2}, \textbf{Y}_{t2}) + s(\textbf{X}_{t3}, \textbf{Y}_{t3}). \label{multi-t emd}
\end{equation} 
\noindent Using Eq. (\ref{multi-t emd}), semantic relevance between two skeleton sequences is measured at multiple temporal scales, as shown in Fig.~\ref{fig:cross optimal match} (b) (highlighted in blue color).  

\noindent\textbf{Cross-Scale Matching.}
As discussed in Sec. \ref{intro}, different instances of the same action class may be performed at different magnitudes (spatial scales). 
For the class, such as `hand waving', people may perform it by moving their palm only (joint-level), or by moving the forearm (part-level), or even by moving the whole arm (limb-level).
Additionally, the same-category samples can also be performed at different speeds (temporal scale).
Thus, there is also semantic relevance between different scales' skeleton representations for matching.
To address the cross-scale matching challenge, we further investigate how to measure the semantic relevance between skeleton sequences across different scales, including cross-spatial scale and cross-temporal scale matching, considering the possibility of different spatial magnitudes and temporal speeds of the same action.

For cross-spatial scale matching, the three spatial scales' skeleton representations are  $\textbf{X}_{s1} \in \mathbbm{R}^{C \times N \times T}$, $\textbf{X}_{s2} \in \mathbbm{R}^{C \times N_2 \times T}$, and $\textbf{X}_{s3} \in \mathbbm{R}^{C \times N_3 \times T}$.
It can be seen that all these three scales' representations contain $T$ frame features.
Thus, we first perform 1D average pooling ($AvgPool$) on the spatial dimension to generate these three scales' features in the same shape ($\mathbbm{R}^{C \times T}$), and then formulate the semantic relevance score as the $T$ frame features optimal matching problem to acquire the optimal matching flow between different spatial-scale representations.
The cross-spatial scale semantic relevance score can be represented as:
\begin{equation}
\small
    s_{cs}(\textbf{X}, \textbf{Y}) = \sum_{i=1}^{3}\sum_{j=1, j \neq i}^{3} s(AvgPool(\textbf{X}_{si}), AvgPool(\textbf{Y}_{sj})). \label{cross-s emd}
\end{equation}
\noindent This process is shown in Fig.~\ref{fig:cross optimal match} (a) (highlighted in orange color).
In this way, we solve the problem of matching skeleton sequences with different motion magnitudes.

Furthermore, we also address the problem of matching the skeleton sequences with different motion speeds and design the cross-temporal scale matching.
The three temporal scales' skeleton features are represented as $\textbf{X}_{t1} \in \mathbbm{R}^{C \times N \times T}$, $\textbf{X}_{t2} \in \mathbbm{R}^{C \times N \times T/2}$, and $\textbf{X}_{t3} \in \mathbbm{R}^{C \times N \times T/4}$. 
It can be seen that these three scales' representations all contain $N$ joint features.
Similarly, the 1D average pooling ($AvgPool$) can be performed on the temporal dimension to pool $\textbf{X}_{t1}$, $\textbf{X}_{t2}$, and $\textbf{X}_{t3}$ into the same shape ($\mathbbm{R}^{C \times N}$). 
Thus, the Earth Mover's Distance is used to measure the semantic relevance score across different temporal scales as:
\begin{equation}
\small
    s_{ct}(\textbf{X}, \textbf{Y}) = \sum_{i=1}^{3}\sum_{j=1, j \neq i}^{3} s(AvgPool(\textbf{X}_{ti}), AvgPool(\textbf{Y}_{tj})),  \label{cross-t emd}
\end{equation}
\noindent This process is shown in Fig.~\ref{fig:cross optimal match} (b) (highlighted in orange color).
Action sequences with different motion speeds can be matched well through Eq. (\ref{cross-t emd}).

\noindent\textbf{Summary.}
As above mentioned, we first introduce the single-scale semantic relevance score  $s(\textbf{X}, \textbf{Y})$ (Eq. (\ref{single-scale})) that considers the useful spatial structure and temporal order information during matching.
To address the problem that different instances of the same action class samples may be performed at different paces with different motion dynamics, we introduce 4 types of semantic relevance scores including multi-spatial scale ($s_{ms}(\textbf{X}, \textbf{Y})$, Eq. (\ref{multi-s emd})), multi-temporal scale ($s_{mt}(\textbf{X}, \textbf{Y})$, Eq. (\ref{multi-t emd})), cross-spatial scale ($s_{cs} (\textbf{X}, \textbf{Y})$, Eq. (\ref{cross-s emd})), and cross-temporal scale ($s_{ct}(\textbf{X}, \textbf{Y})$, Eq. (\ref{cross-t emd})). 
The semantic relevance score for our proposed model is averaging from multi-scale and cross-scale relevance scores, which is then used to predict the action category.

\subsection{Objective Loss, Model Training, and Inference}
{
The majority of current few-shot learning techniques~\cite{simon2020adaptive,snell2017prototypical, xu2021learning,ye2020fewshot,Zhang_2020_CVPR} implement a pre-training stage prior to meta-learning. The effectiveness of this stage within the realm of few-shot learning has been validated by~\cite{chen2018a, chen2021meta}.
}
Therefore, the proposed method is trained in two sequential stages: The \textit{First} is the pre-training stage. The embedding network is trained on \textit{meta-training set $D_{train}$} in a standard supervised learning way (Sec. \ref{Pretraining}).
The \textit{Second} is the \textit{meta-training} stage. 
{The embedding network and our optimal matching method (Sec. \ref{Matching}) are further optimized in an end-to-end manner, following~\cite{rubner2000earth,zhang2020deepemd}. 
Both training stages leverage the Softmax cross-entropy loss as the classification loss $L_{cls}$ for optimization (we experimented with the normalized Softmax cross-entropy loss and Angular margins as objective functions, and the Softmax cross-entropy loss yielded the best performance). }
Given an unseen query sequence $q$ and its support set $S$ (both sampled from \textit{meta-testing set $D_{test}$}) at test, the goal is to determine which support set classes $q$ belongs to.

\vspace{-0.3cm}
\section{Experiments}
\subsection{Datasets}

\noindent{\bf NTU RGB+D} dataset \cite{ntu60} consists of 56880 skeleton action sequences, which is the most widely-used dataset in skeleton-based action recognition research. In this dataset, action samples are performed by 40 subjects, with three camera views, and categorized into 60 classes.
The NTU RGB+D dataset provides two standard evaluation protocols, namely cross-view (CV) and cross-subject (CS).

\noindent{\bf NTU RGB+D 120} dataset \cite{ntu120} is currently the largest dataset with 3D joints annotations for human action recognition. The dataset contains 114480 action samples in 120 action classes. Action samples are captured by 106 volunteers with three camera views. This dataset contains 32 setups, and each setup denotes a specific location and background. 
The evaluation protocols of this dataset are (1) cross-subject evaluation and (2) cross-setup evaluation.

\noindent{\bf PKU-MMD} dataset \cite{pkummd} is a large-scale benchmark for continuous multi-modality 3D skeleton action understanding.
It contains 21,545 action instances performed by 66 distinct subjects in 51 action categories.
The dataset also utilizes CV and CS evaluation protocols.
PKU-MMD consists of two subsets, part I and part II, we conduct experiments on the part I subset in this work.

\vspace{-0.3cm}
\subsection{Training and Evaluation Protocol}

\noindent\textbf{Training Protocol.} 
As described in Sec. \ref{ProblemFormulation}, we formulate the one-shot skeleton action recognition problem as a meta-learning problem~\cite{snell2017prototypical,vinyals2016matching,Zhang_2020_CVPR,cao2020few,ECCV2020fewshot}. Therefore, we adopt the meta-training phase as our training protocol.
All experiments are trained under the \textit{5}-way, \textit{1}-shot setting, and we set the number of query samples for each class to 15.

\noindent\textbf{Evaluation Protocol 1.}
Similarly, we conduct the evaluation on \textit{5}-way, \textit{1}-shot setting, and adopt this procedure as Evaluation Protocol 1 in the following experiments.

\noindent\textbf{Evaluation Protocol 2.} 
To compare with the existing one-shot skeleton action recognition techniques~\cite{ntu120,memmesheimer2020skeleton,memmesheimer2020signal, sabater2021one,wang2022temporal,wang2022uncertainty} in a fair way, we also follow the official one-shot protocol described in~\cite{ntu120} for the dataset NTU RGB+D 120.
Specifically, the testing set consists of 20 novel classes, and we pick one sample from each novel class as the exemplar\footnote{More exemplars \& dataset splitting details are provided in the appendix. \label{exemplar}} and leave the rest (except for the 20 exemplars) to test the recognition performance. 
For datasets NTU RGB+D and PKU-MMD\textsuperscript{\ref{exemplar}}, we adopt a similar protocol in experiments. 

\subsection{Implementation Details}
\noindent\textbf{Pre-training Stage.}
We use the SGD optimizer with Nesterov momentum (0.9) as the optimizer.
The learning rate is set as 0.1 and is divided by 10 at $30_{th}$ epoch and $40_{th}$ epoch. 
The training process is ended at the $50_{th}$ epoch.
For NTU RGB+D and NTU RGB+D 120, the batch size is set as 64. The batch size for the PKU-MMD dataset is 32.

\noindent\textbf{Meta-learning Stage.}
the learning rate starts at 0.001 and decays every 10 epochs by 0.5.
We train for 100 epochs using SGD optimizer, and each epoch consists of 100 episodes from the \textit{meta-training set $D_{train}$}. For the meta-testing phase, we sample 1000 episodes from \textit{meta-testing set $D_{test}$}. 
($D_{train}$ and $D_{test}$ are defined in Sec. \ref{ProblemFormulation}).

\noindent\textbf{Embedding Network.} 
We use Adaptive Graph Convolutional Network \cite{Shi_2019_CVPR_twostream} (AGCN) as our single-scale embedding network which has 9 GCN blocks.
For multi-spatial and multi-temporal networks, the first 6 blocks are shared to capture the single-scale features. Then each scale feature is processed by the other 3 individually and in parallel\footnote{The detailed network structures can be found in appendix.\label{network}}.

\subsection{Dataset Splitting}
For dataset NTU RGB+D 120, we adopt the one-shot skeleton action setting as described in~\cite{ntu120} which splits the full dataset into a training set and a testing set. The action classes of the two sets are distinct which include 100 classes for training and 20 for testing\textsuperscript{\ref{exemplar}}. For dataset NTU RGB+D, the training set and testing set are determined by the selection of 50 classes and 10 classes from the 100 training and 20 testing classes of the NTU RGB+D 120, respectively. Similarly, for dataset PKU-MMD, we divide the action categories into a training set and a testing set which include 41 classes for training and 10 classes for testing\textsuperscript{\ref{exemplar}}.

As no hold-out validation set is defined in the one-shot skeleton action setting and all these three datasets contain the cross-subject setting for supervised action recognition. Therefore, we divide the training class data into the training set and validation set based on the cross-subject principle for the one-shot skeleton action recognition task.
We maintain the testing set of these three datasets as the testing set for one-shot skeleton action recognition.

\begin{table}[t]
  \centering
  \caption{One-shot skeleton recognition experiment under the Evaluation Protocol 1. (S-scale: single-scale matching; M-scale: multi-scale matching; M\&C-scale: multi-scale and cross-scale matching)
  }
  \vspace{-0.2cm}
  \begin{tabular}{l|c|c|c}
    \hline
    Method & NTU & NTU 120 & PKU-MMD\\
    \hline
    ProtoNet~\cite{snell2017prototypical} & 78.3 & 80.3 & 84.7 \\
    FEAT~\cite{ye2020fewshot} & 77.8 & 80.0 & 83.8 \\
    Subspace~\cite{simon2020adaptive} & 77.9 & 80.5 & 84.2 \\
    Dynamic Filter~\cite{xu2021learning} & 79.3 & 80.4 & 84.9 \\
    \hline
    S-scale (Ours) &  80.4 & 81.2 & 85.7 \\
    M-scale (Ours) & 82.6  & 83.5 & 88.2 \\
    M\&C-scale (Ours) & \textbf{83.7}  & \textbf{84.5}  & \textbf{89.3} \\
    \hline
  \end{tabular}
  \label{tab:protocol 1}
  \vspace{-0.5cm}
\end{table}

\subsection{Evaluating on One-Shot Skeleton Action Recognition}
We conduct extensive experiments with five optimal matching strategies that include single-scale (`\textbf{S-scale}'), multi-spatial scale (`\bm{$M_s$}'), multi-temporal scale (`\bm{$M_t$}'), cross-spatial scale (`\bm{$C_s$}'), and cross-temporal scale (`\bm{$C_t$}'). We combine
`\bm{$M_s$}' and `\bm{$M_t$}' to form a new multi-scale strategy `\textbf{M-scale}'. In addition, we combine `\bm{$M_s$}', `\bm{$M_t$}', `\bm{$C_s$}', and `\bm{$C_t$}' to form another new strategy `\textbf{M\&C-scale}' that matches spatial and temporal features simultaneously at multiple scales and also cross scales.

We compare our method with two groups of state-of-the-art methods on one-shot skeleton action recognition. The first group consists of state-of-the-art few-shot image classification methods including Subspace~\cite{simon2020adaptive}, ProtoNet~\cite{snell2017prototypical}, Dynamic Filter~\cite{xu2021learning}, FEAT~\cite{ye2020fewshot}. All these methods use the same embedding network as our method for fair comparisons. We re-implement~\cite{simon2020adaptive,snell2017prototypical,xu2021learning,ye2020fewshot} based on publicly available codes and conduct experiments on NTU RGB+D, NTU RGB+D 120 and PKU-MMD datasets. The implementation details are available in the appendix. 
The second group consists of state-of-the-art one-shot skeleton action recognition techniques including APSR~\cite{ntu120}, TCN~\cite{sabater2021one}, SL-DML~\cite{memmesheimer2020signal}, Skeleton-DML~\cite{memmesheimer2020skeleton}, uDTW~\cite{wang2022uncertainty}, and JEANIE~\cite{wang2022temporal}. 
For those one-shot skeleton action recognition works~\cite{ntu120,memmesheimer2020skeleton,memmesheimer2020signal,sabater2021one,wang2022temporal,wang2022uncertainty}, the results in Tabs.~\ref{tab:protocol 1},\ref{tab:proto 2},\ref{tab:abla classes} are from the original papers.
We compare our method with the first group methods under both evaluation protocols, while the second group with Evaluation Protocol 2 only. 

\begin{table}[t]
  \caption{One-shot skeleton recognition experiments under the Evaluation Protocol 2. (S-scale: single-scale matching; M-scale: multi-scale matching; M\&C-scale: multi-scale and cross-scale matching)
  }
  \vspace{-0.2cm}
  \centering
  \begin{tabular}{l|c|c|c}
    \hline
    Method & NTU & NTU 120 & PKU-MMD\\
    \hline
    Attention Network~\cite{liu2017global} &-- & 41.0 & --\\
    Fully Connected~\cite{liu2017global} & --& 42.1 &--\\
    Average Pooling~\cite{liu2017skeleton} & -- & 42.9 &--\\
    APSR~\cite{ntu120}  & --& 45.3 &--\\
    TCN~\cite{sabater2021one} & -- & 46.5 &--\\

    SL-DML~\cite{memmesheimer2020signal} &-- & 50.9 &--\\
    Skeleton-DML~\cite{memmesheimer2020skeleton} &  --& 54.2 &--\\
    uDTW~\cite{wang2022uncertainty} & 72.4 &  49.0 &--\\
    JEANIE~\cite{wang2022temporal} & 80.0 &  57.0 &--\\
    ALCA-GCN~\cite{Zhu_2023_WACV} & -- &  57.6 &--\\
    
    \hline
    ProtoNet~\cite{snell2017prototypical} & 74.8 & 60.4 & 78.1 \\
    FEAT~\cite{ye2020fewshot} & 74.3 & 61.5 & 75.9 \\
    Subspace~\cite{simon2020adaptive} & 75.6 & 60.9 & 75.6 \\
    Dynamic Filter~\cite{xu2021learning} & 75.9 & 60.6 & 78.8\\
    \hline
    S-scale (Ours)  & 77.4  & 63.2 & 82.8\\
    M-scale (Ours) & 81.6  &  67.6 & 85.7 \\
    M\&C-scale (Ours) & \textbf{82.7}  & \textbf{68.7}  &  \textbf{86.9}\\
    \hline
  \end{tabular}
  \label{tab:proto 2}
  \vspace{-0.5cm}
\end{table}

\subsubsection{Evaluation Protocol 1:}
We evaluate the \textit{5}-way \textit{1}-shot setting on all three datasets and Tab.~\ref{tab:protocol 1} shows experimental results. It can be seen that our proposed single-scale optimal matching outperforms state-of-the-art few-shot learning methods on all three datasets. In addition, our proposed multi-scale and cross-scale matching strategies further improve one-shot skeleton action recognition by large margins, 
demonstrating the effectiveness of our proposed method on the one-shot skeleton action recognition task.

\subsubsection{Evaluation Protocol 2:}
Following the one-shot setting in~\cite{ntu120}, we also conduct experiments on NTU RGB+D, NTU RGB+D 120, and PKU-MMD datasets under Evaluation Protocol 2. Similar to the experiments on Evaluation Protocol 1, our proposed method outperforms the state-of-the-art one-shot skeleton action recognition and few-shot learning methods by large margins (up to $8\%$ on NTU120 and PKU-MMD). Tab.~\ref{tab:proto 2} shows more details of the experiments.

\subsection{Ablation Studies}
\noindent\textbf{Matching Strategies:} We compare our proposed optimal matching with the \textit{global matching scheme} that adopts global average pooling to generate feature vector, as well as \textit{a local matching scheme} that computes joint-to-joint distances (either Euclidean or Cosine distance) with local-level representations. For fair comparisons, we adopt the same backbone and training scheme for all compared methods, and Tab.~\ref{tab:abla fv fm emd} reports experiment results. We can observe that models with local-level representations
perform better than models that rely on global-level representations (in globally pooled feature vectors).  In addition, our method which 
works with the optimal matching flow between all pairs of joints outperforms all compared methods. Note that all experiments were conducted on the `S-scale' model under evaluation protocol 2. 

\begin{table}[t]
  \centering
  \caption{
  Comparison of different feature embedding approaches and distance metrics under Evaluation Protocol 2.
  (EMD: earth mover's distance)}
  \vspace{-0.2cm}
  \begin{tabular}{l|c|c|c|c}
    \hline
    Embedding & Metric & NTU &  NTU 120 & PKU-MMD \\
    \hline
    Global  & Euclidean & 71.6 & 56.1   & 78.1\\
    Global  & Cosine    & 74.8 & 60.4   & 74.8\\
    \hline
    Local   & Euclidean & 74.7  & 58.0   & 80.7 \\
    Local   & Cosine    & 75.8  & 61.3   & 80.8 \\
    Local   & EMD       & \textbf{77.4}  &  \textbf{63.2}   & \textbf{82.8} \\
    \hline
  \end{tabular}
  \label{tab:abla fv fm emd}
  \vspace{-0.2cm}
\end{table}

\begin{table}[t]
  \centering
  \caption{Experiments on different sizes of the auxiliary training set for one-shot skeleton recognition on NTU RGB+D 120 dataset. (S-scale: single-scale matching; M-scale: multi-scale matching; M\&C-scale: multi-scale and cross-scale matching)
  }
  \vspace{-0.2cm}
  \begin{tabular}{l|c|c|c|c|c}
    \hline
    Train Classes &  20 & 40 & 60 & 80 &  100\\
    \hline
    APSR~\cite{ntu120} & 29.1 & 34.8 & 39.2 & 42.8 & 45.3  \\
    SL-DML~\cite{memmesheimer2020signal} & 36.7 & 42.4 & 49.0 & 46.4 & 50.9 \\
    Skeleton-DML~\cite{memmesheimer2020skeleton} & 28.6 & 37.5 & 48.6 & 48.0 & 54.2 \\
    uDTW~\cite{wang2022uncertainty} & 32.2 & 39.0 & 41.2 & 45.3 & 49.0 \\
    JEANIE~\cite{wang2022temporal} & 38.5 & 44.1 & 50.3 & 51.2 & 57.0 \\
    ALCA-GCN~\cite{Zhu_2023_WACV} & 38.7 & 46.6 & 51.0 & 53.7 & 57.6 \\
    \hline
    ProtoNet~\cite{snell2017prototypical} & 36.7 & 45.7 & 54.4  & 55.5 & 60.4 \\
    FEAT~\cite{ye2020fewshot} & 37.6 & 44.2 & 52.2 & 55.4 & 61.5 \\
    Subspace~\cite{simon2020adaptive} & 37.7 & 45.3 & 54.2 & 55.5 & 60.9 \\
    Dynamic Filter~\cite{xu2021learning} & 34.5 & 43.0 & 54.0 & 52.9 & 60.6 \\
    \hline
    S-scale (Ours) & 37.9 & 46.5 & 54.6 & 58.7 & 63.2  \\
    M-scale (Ours) &  41.2 & 52.6  & 59.0 & 62.4 & 67.6 \\
    M\&C-scale (Ours) & \textbf{44.1}  & \textbf{55.3}  & \textbf{60.3} & \textbf{64.2} & \textbf{68.7} \\
    \hline
  \end{tabular}
  \label{tab:abla classes}
  \vspace{-0.3cm}
\end{table}

\noindent\textbf{Reducing Training Classes:} While evaluating one-shot action recognition methods, one interesting question is how many training classes are required to achieve fair recognition performance. We examine this issue under Evaluation Protocol 2 by following
prior studies on NTU RGB+D 120 dataset \cite{ntu120,memmesheimer2020skeleton,memmesheimer2020signal,wang2022temporal,wang2022uncertainty}. Tab. \ref{tab:abla classes} shows experimental results. It can be seen that our method outperforms the state-of-the-art by large margins under different numbers of training classes.
With a training set of 60 classes, our method
is on par with state-of-the-art methods that are trained by using 100-class training set. This clearly shows the effectiveness of our proposed optimal matching.

\noindent\textbf{Effect of Multi-Scale and Cross-Scale Matching Manners:}
We conduct experiments on different combinations of our proposed optimal matching strategies. Tab. \ref{tab:Abla multi scale cross scale} shows experimental results under evaluation protocol 2. We can see that including any of our proposed matching strategies ($M_s$, $M_t$, $C_s$, and $C_t$) improves the one-shot skeleton action recognition clearly. Including all four matching strategies perform simply the best over all three datasets, demonstrating the effectiveness of our proposed optimal matching technique.
Please refer to the appendix for more ablation studies and visualization of the proposed multi-scale and cross-scale optimal matching strategies. 

\begin{table}[t]
  \caption{Evaluation of different combinations of optimal matching manners under the Evaluation Protocol 2: The second line shows the baseline result under the single-scale matching manner. ($M_t$: multi-temporal scale matching; $M_s$: multi-spatial scale matching;
  $C_t$: cross-temporal scale matching;
  $C_s$: cross-spatial scale matching)
  }
  \vspace{-0.2cm}
  \centering
  \begin{tabular}{c|c|c|c|c|c|c}
    \hline
     $M_t$ & $M_s$ & $C_t$ & $C_s$ & NTU & NTU 120 & PKU-MMD\\
    \hline
      &  &  &  & 77.4  & 63.2 & 82.8\\
    \checkmark & & & & 79.3 & 65.0 & 85.0\\
    &  \checkmark & & & 79.6 & 65.1 & 84.9\\
    \checkmark &  \checkmark & & & 81.6  &  67.6 & 85.7 \\
    \checkmark & & \checkmark & & 80.7 & 66.7 & 85.7\\
    & \checkmark & & \checkmark & 80.8 & 67.3 & 86.0\\
     \checkmark & \checkmark & \checkmark & \checkmark & \textbf{82.7}  & \textbf{68.7}  &  \textbf{86.9}\\
    \hline
  \end{tabular}
  \vspace{-0.2cm}
  \label{tab:Abla multi scale cross scale}
\end{table}

\noindent\textbf{Effect of Multiple Spatial Scales and Multiple Temporal Scales: } We also study how models with different spatial and temporal scales (as the backbone) perform for the one-shot skeleton action recognition task. Tab. \ref{tab:abla multi scale} shows the one-shot learning performance with different combinations of scales. We can observe that the
model performs best when combining 
scales 1, 2, and 3. In addition, employing two scales (scales 1 and 2 or scales 1 and 3) also outperform the model using scale 1 only, showing the benefits of the proposed multi-scale representations. 

\noindent {More ablation studies about temporal pooling strategies, feature pooling strategies and feature extractors can be found in the Appendix.} 

\begin{table}[t]
  \centering
  \caption{
  Evaluation of our proposed multi-scale matching with different scale combinations. The experiments were conducted on NTU RGB+D 120 dataset under the Evaluation Protocol 2.}
  \vspace{-0.2cm}
  \begin{tabular}{l|c|c}
    \hline
    scales & Multi-Spatial Scale & Multi-Temporal Scale \\
    \hline
    1      &  63.2 & 63.2 \\
    1,2    &  64.3 & 64.4 \\
    1,3    &  64.6 & 64.2 \\
    1,2,3  &  \textbf{65.1} & \textbf{65.0} \\
    \hline
  \end{tabular}
  \label{tab:abla multi scale}
  \vspace{-0.5cm}
\end{table}

\section{Conclusion and Future Works}
In this paper, we address the one-shot skeleton action recognition as a matching problem.
We obtain the multi-spatial and multi-temporal scale features by designing a hierarchical pooling that represents the same skeleton sequence at various spatial and temporal scales.
Moreover, based on the multi-scale skeleton features, we propose a multi-scale skeleton matching strategy and a cross-scale skeleton matching manner to measure the semantic relevance between two skeleton sequences for one-shot skeleton action recognition.
The experiments demonstrate that our proposed method achieves superior one-shot skeleton action recognition performance.
{At present, our proposed method is applied specifically to the Kinect V2 skeleton format. In our upcoming research, we plan to broaden our scope to include one-shot skeleton action recognition using various skeleton formats, possibly even exploring recognition across different formats. Furthermore, we also consider integrating one-shot skeleton action recognition with the large language models (LLMs), leveraging their vast reservoir of prior knowledge.}

{
\small
\bibliographystyle{IEEEtran}
\bibliography{egbib}
}

\appendices
\section{Embedding Network Structures}
\begin{figure}[t]
\begin{center}
\includegraphics[trim=0cm 0cm 0cm 0cm,clip, width=0.25\textwidth]{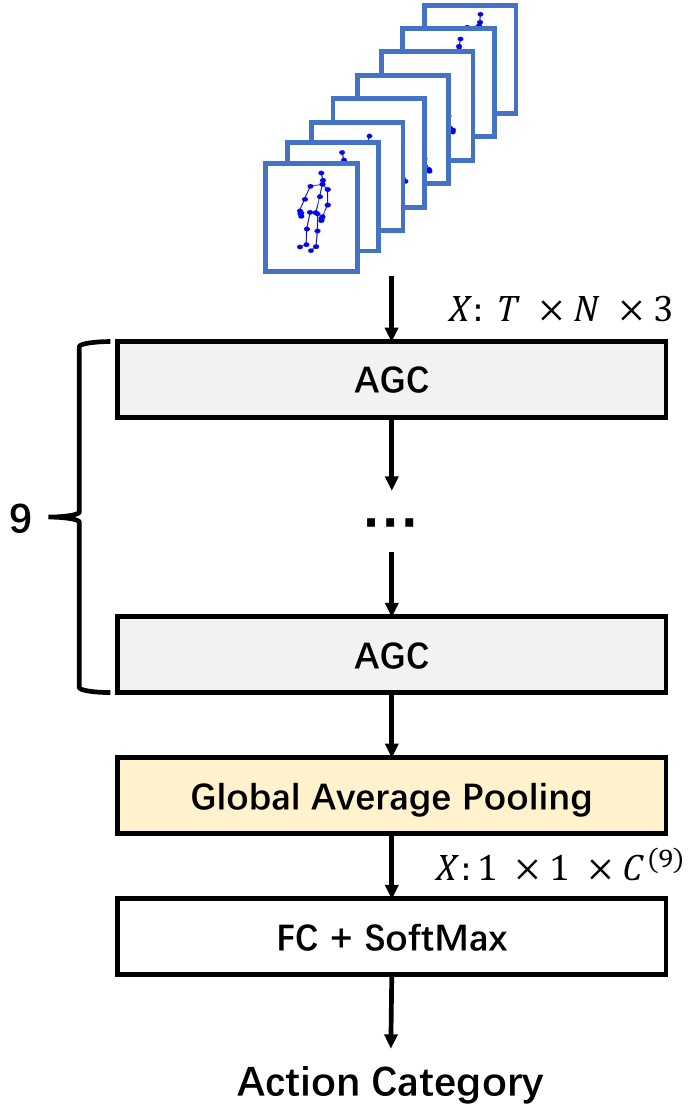}
\end{center}
  \caption{
  Illustration of the single-scale embedding network (AGCN \cite{Shi_2019_CVPR_twostream}). There are a total of 9 AGC blocks, followed by a global average layer and a \textit{softmax} classifier. 
  (\textit{N} denotes the number of skeleton joints, \textit{T} denotes the number of frames, and $C^{(i)}$ denotes the number of output channels at $i^{th}$ block.)
  }
  \vspace{-0.5cm}
\label{fig:single-scale}
\end{figure}

\noindent\textbf{Single scale.}
We adopt the adaptive graph convolutional network (AGCN) \cite{Shi_2019_CVPR_twostream} as our single scale skeleton embedding network, which is the stack of 9 adaptive graph convolutional (AGC) blocks, global average pooling layer, and a \textit{softmax} classifier, as shown in Fig. \ref{fig:single-scale}.
For the details of the AGC block and the setting of each block, please refer to \cite{Shi_2019_CVPR_twostream}.

\begin{figure*}[htb]
\begin{center}
\includegraphics[trim=0.2cm 0.2cm 0.2cm 0.1cm,clip, width=0.75\textwidth]{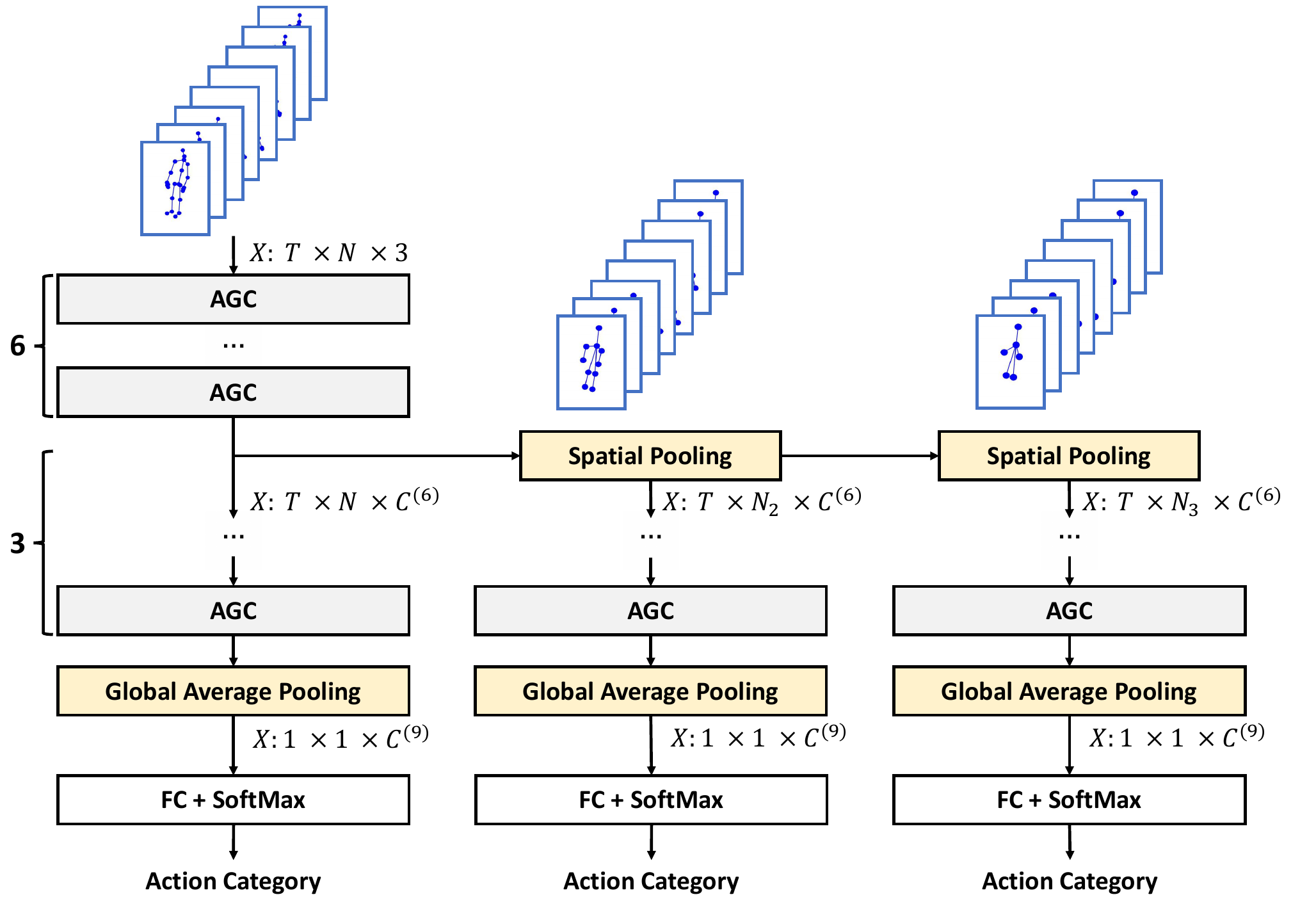}
\end{center}
\vspace{-0.3cm}
  \caption{
  Illustration of multi-spatial scale embedding network. After processing the first six AGC blocks, we perform spatial pooling to generate the spatial scale 2 and spatial scale 3 features. Finally, the multi-spatial scale features are trained individually and paralleled. For the details of spatial pooling, please refer to Fig. (2)(a) in the main manuscript.
  (\textit{N} denotes the number of skeleton joints, $N_2$ denotes the number of nodes for the second-scale spatial graph, $N_3$ stands for the number of third-scale
  spatial graph nodes, \textit{T} denotes the number of frames, and $C^{(i)}$ denotes the number of output channels at $i^{th}$ block.)
  }
  \vspace{-0.3cm}
\label{fig:multi-spatial scale}
\end{figure*}

\noindent\textbf{Multi-spatial scale.}
To construct the multi-spatial scale skeleton, we first build six AGC blocks on the original spatial scale to capture the joint-wise feature representation and then perform the spatial pooling to generate the spatial scale 2 and spatial scale 3 features.
{
To extract the multi-spatial scale skeleton representations, 
all three spatial streams undertake classification during the pre-training stage, as shown in Fig. \ref{fig:multi-spatial scale}. Each stream is independently optimized using the cross-entropy loss.
}
The vertically parallel AGC blocks share the same setting, except for the skeleton graph structure.

\begin{figure*}[htb]
\begin{center}
\includegraphics[trim=0.2cm 0.2cm 0.2cm 0.1cm,clip, width=0.75\textwidth]{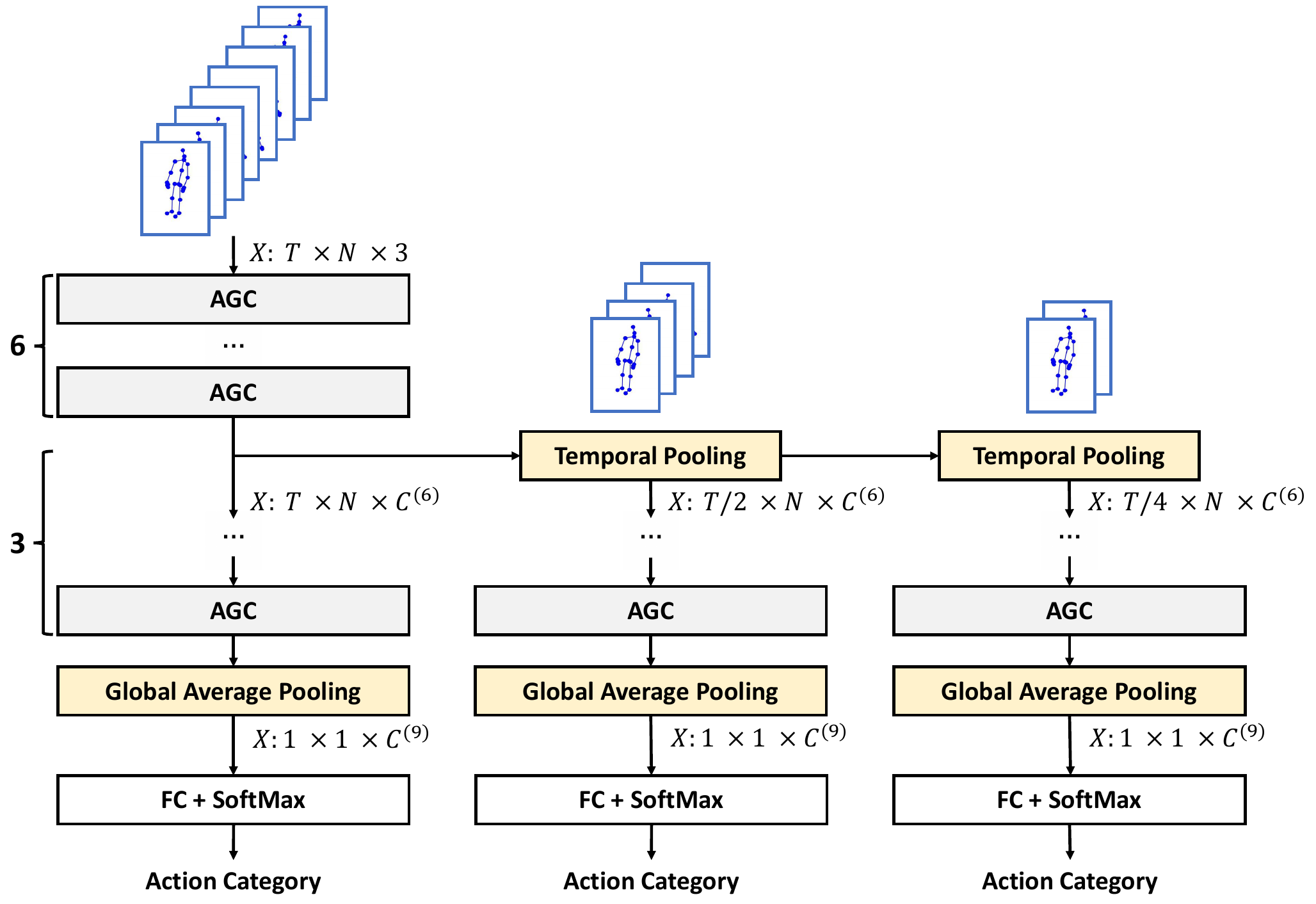}
\end{center}
\vspace{-0.2cm}
  \caption{
  Illustration of multi-temporal scale embedding network.
  First, we build six AGC blocks on the original temporal scale to capture the feature representation and then perform the temporal pooling to generate the coarser scales' features.
  please refer to Fig. (2)(b) in the main manuscript for the details of temporal pooling.
  (\textit{N} denotes the number of skeleton joints, \textit{T} denotes the number of frames, and $C^{(i)}$ denotes the number of output channels at $i^{th}$ block.)
  }
\vspace{-0.2cm}
\label{fig:multi-temporal scale}
\end{figure*}

\noindent\textbf{Multi-temporal scale.}
Similarly, after processing the first six AGC blocks in the original temporal scale, we perform the temporal pooling to generate the coarser scales (temporal scale 2 and temporal scale 3).
{
To extract the multi-temporal scale skeleton representations, each of the three temporal streams is subject to classification during the pre-training stage and is optimized individually using the cross-entropy loss, as shown in Fig. \ref{fig:multi-temporal scale}.
}
The vertically parallel AGC blocks share the same setting.

For all types of embedding networks (single scale, multi-spatial scale, and multi-temporal scale), we leverage the output feature maps of $9^{th}$ block as the skeleton feature representations in the \textit{meta-training} stage.

\section{Additional experimental Results}
\label{results}

\noindent\textbf{More experimental results on different combinations of optimal matching manners:}
We conduct more experiments on different combinations of optimal matching manners to show the advantages of our proposed optimal matching strategies.
Tabs. \ref{tab:protocol 1 supp} and \ref{tab:abla classes supp} show the experimental results under the evaluation protocol 1 and the ``Reducing training classes'' experiments, respectively. Similar to the conclusion we got in the main manuscript, we can see that including any of our proposed matching strategies ($M_s$, $M_t$, $C_s$, and $C_t$) improves the one-shot skeleton action recognition performance clearly. Including all four matching strategies perform the best on all three datasets, demonstrating the effectiveness of our proposed optimal matching manners.

\begin{figure*}[htb]
\begin{center}
\includegraphics[trim=0cm 0cm 0cm 0cm,clip, width=0.85\textwidth]{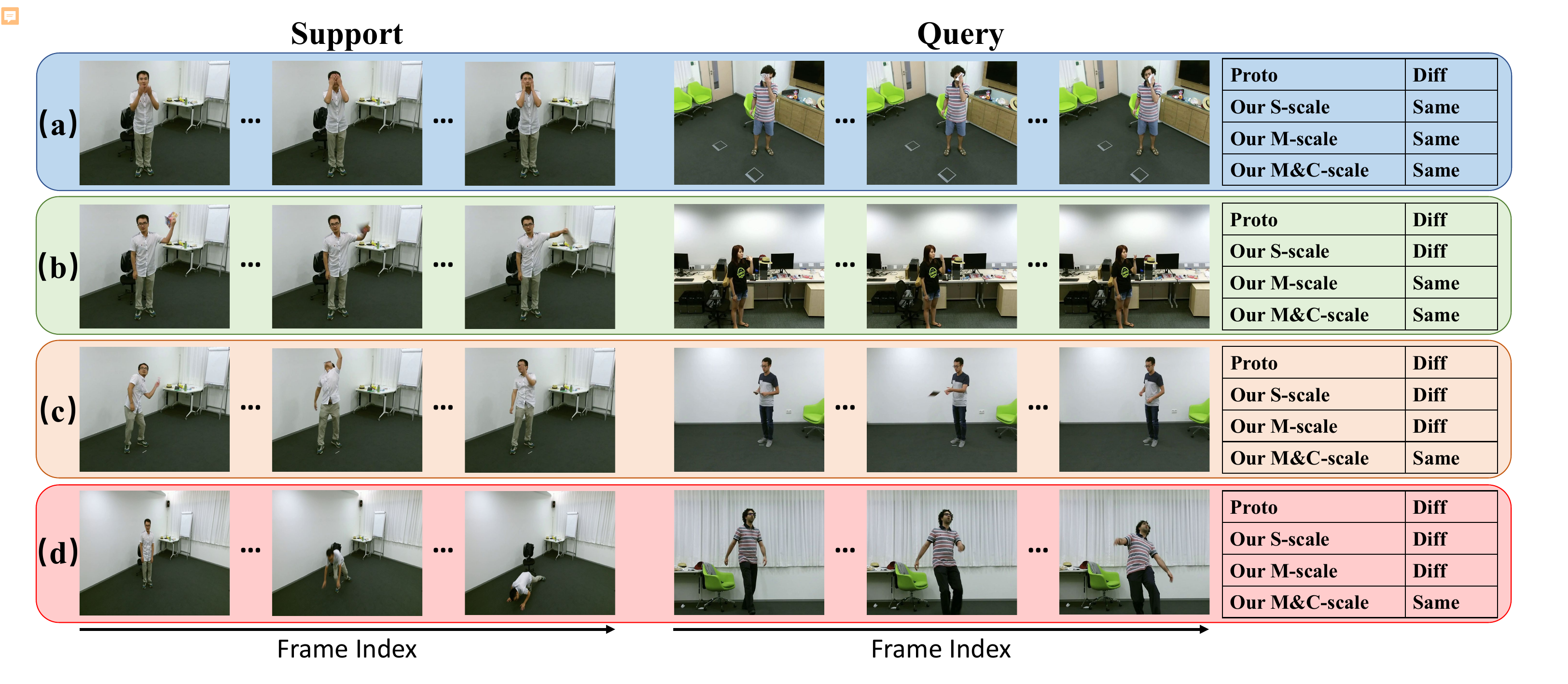}
\end{center}
  \caption{
 Visualization of comparison results between different matching strategies.
 For better visualization, we here show the RGB videos, instead of skeleton sequences.
 (`Same' means that model predicts the support and query samples are from the same action category; `Diff' stands for that the predicted action classes for support and query samples are different.)
  } 
  \vspace{-0.4cm}
\label{fig:visualization}
\end{figure*}

\begin{table*}
\parbox{.49\linewidth}{
  \centering
  \caption{
  Evaluation of different combinations of optimal matching manners under Evaluation Protocol 1. The second line shows the baseline result under the single-scale matching manner. ($M_t$: multi-temporal matching; $M_s$: multi-spatial matching;
  $C_t$: cross-temporal matching;
  $C_s$: cross-spatial matching)
  }
  \vspace{-0.2cm}
   \resizebox{.47\textwidth}{!}{
  \begin{tabular}{c|c|c|c|c|c|c}
    \hline
    $M_t$ & $M_s$ & $C_t$ & $C_s$ & NTU & NTU 120 & PKU-MMD\\
    \hline
    &  &  &  &  80.4 & 81.2 & 85.7 \\
   \checkmark & & & & 81.5 & 82.4 & 87.4 \\
    &  \checkmark & &  & 81.6  & 82.3 & 87.5 \\
    \checkmark &  \checkmark & && 82.6  & 83.5 & 88.2 \\
   \checkmark & & \checkmark &  & 82.9 & 83.5 &  88.3\\
    & \checkmark & & \checkmark & 83.0 & 83.7& 88.4 \\
    \checkmark & \checkmark & \checkmark & \checkmark & \textbf{83.7}  & \textbf{84.5}  & \textbf{89.3} \\
    \hline
  \end{tabular}
  \label{tab:protocol 1 supp}
}
}
\hfill
\parbox{.49\linewidth}{
  \centering
  \caption{Evaluation of different combinations of matching manners on the ``Reducing training classes'' ablation study. The second line shows the baseline result under the single-scale matching manner. ($M_t$: multi-temporal matching; $M_s$: multi-spatial matching;
  $C_t$: cross-temporal matching;
  $C_s$: cross-spatial matching)
  }
  \vspace{-0.2cm}
 \resizebox{.47\textwidth}{!}{
  \begin{tabular}{c|c|c|c|c|c|c|c|c}
    \hline
    $M_t$ & $M_s$ & $C_t$ & $C_s$  &  20 & 40 & 60 & 80 &  100\\
    \hline
    &  &  &  & 37.6 & 46.5 & 54.2 & 58.7 & 63.2  \\
    \checkmark & & & &  39.0 & 50.5 & 55.1 & 60.0 & 65.0  \\
    &  \checkmark & &  & 40.2 & 50.4 & 57.8 & 60.2 & 65.1  \\
    \checkmark &  \checkmark & &&  41.2 & 52.6  & 59.0 & 62.4 & 67.6 \\
    \checkmark & & \checkmark &  & 40.2 & 53.3 & 56.5 &  63.5 &  66.7\\
     & \checkmark & & \checkmark & 42.1 & 51.7 & 58.9 & 61.5 & 67.3 \\
   \checkmark & \checkmark & \checkmark & \checkmark & \textbf{44.1}  & \textbf{55.3}  & \textbf{60.3} & \textbf{64.2} & \textbf{68.7} \\
    \hline
  \end{tabular}
  \label{tab:abla classes supp}
}
}
\vspace{-0.2cm}
\end{table*}

\noindent {\textbf{More experimental results on different numbers of temporal scales:}
We conduct additional experiments on different numbers of temporal scales to show the effectiveness of the proposed multi-temporal scale strategy. 
Specifically, we conduct experiments on two temporal scales ($T$ and $T/2$) and four temporal scales ($T$, $T/2$, $T/4$, and $T/8$).
The experimental results are shown in Tab.~\ref{tab: temporal scales}. It can be seen that the `4-Scale' result is 64.7, which is a little lower than the `3-Scale' (64.9), showing that using 4 temporal scales is redundant. 
Note that all experiments were conducted on the NTU RGB+D 120 dataset under evaluation
protocol 2.}

\begin{table}[t]
\begin{center}
\caption{Evaluation of our proposed multi-scale temporal matching with different numbers of temporal scales. The experiments were conducted on NTU RGB+D 120
dataset under the Evaluation Protocol 2.
}
\label{tab: temporal scales}
\begin{tabular}{l|c|c|c} 
  \hline 
   Number of scales  & 2-Scale  &  3-Scale   &  4-Scale \\
  \hline
   Multi-Temporal Scale   & 64.4   & \textbf{65.0}  & 64.8 \\
     \hline

\end{tabular}
\end{center}
\end{table}

\noindent {\textbf{More experimental results on different temporal pooling strategies:}
We conduct additional experiments on different temporal pooling strategies. 
Here, we have further expanded our experiments to encompass three additional pooling setups:
(1) pooling performed across spans of 2 and 8 consecutive frames (Pooling2\_8).
(2) pooling performed across spans of 4 and 8 consecutive frames (Pooling4\_8).
(3) pooling performed across spans of 5 and 10 consecutive frames (Pooling5\_10).
The experimental results are shown in Tab.~\ref{tab: temporal poolings}.
We can observe that `Pooling2\_4' performs the best, which is applied the `Pooling2\_4' in our main experiments.
}

\begin{table}[t]
\begin{center}
\caption{Evaluation of our proposed multi-scale temporal matching with different temporal pooling strategies. The experiments were conducted on NTU RGB+D 120
dataset under the Evaluation Protocol 2.
}
\label{tab: temporal poolings}
\resizebox{0.99\linewidth}{!}{
\begin{tabular}{l|c|c|c|c} 
  \hline 
   Temporal Pooling  & Pooling2\_4  &  Pooling2\_8   &  Pooling4\_8 &  Pooling5\_10 \\
  \hline
   Multi-Temporal Scale   & \textbf{65.0}   & 64.7  &64.5 & 64.0 \\
     \hline
\end{tabular}
}
\end{center}
\end{table}

\noindent {\textbf{More experimental results on different feature pooling strategies:}
We conducted this ablation study to evaluate the different feature pooling strategies in cross-scale matching.
In the Earth's Mover Distance, the matching score is computed at the channel level, implying that the channel features must be of the same size. 
In cross-scale matching scenarios, such as cross-temporal scale matching, there's a variation in the size of skeleton features across the three temporal scales. These sizes are denoted as $\textbf{X}_{t1} \in \mathbbm{R}^{C \times N \times T}$, $\textbf{X}_{t2} \in \mathbbm{R}^{C \times N \times T/2}$, and $\textbf{X}_{t3} \in \mathbbm{R}^{C \times N \times T/4}$.
We are considering two potential strategies for addressing this:}

\noindent {(1) \textbf{Pooling All:} Implementing Average Pooling on the temporal dimension to standardize the three temporal scale features to the shape $\mathbbm{R}^{C \times N}$;
}

\noindent {(2) \textbf{Pooling to the min Dimension:} Adjusting the temporal dimension of $\textbf{X}_{t1}$ and $\textbf{X}_{t2}$ through Average Pooling to achieve a size of $4/T$, ensuring compatibility with $\textbf{X}_{t3}$. 
}

{The results from these experiments are presented in Table~\ref{tab: feature pooling}.
We can find that both these two pooling strategies achieve very similar results. Therefore, we opt for the \textbf{Pooling All} strategy, as it offers an optimal balance between performance and efficiency.
}

\begin{table}[t]
\begin{center}
\caption{Evaluation of our cross-scale matching with feature pooling strategies. The experiments were conducted on NTU RGB+D 120
dataset under the Evaluation Protocol 2.
}
\label{tab: feature pooling}
\begin{tabular}{l|c|c} 
  \hline 
    Pooling methods & Pooling All  &  Pooling to the min Dimension   \\
  \hline
  Cross-Temporal scale    & 66.7   & 66.8   \\
  Cross-Spatial scale     & 67.3   & 67.3  \\
  M\&C scale             &  68.7  & 68.8  \\
     \hline

\end{tabular}
\end{center}
\end{table}

\noindent{\textbf{More experimental results on different feature extractors:}
We conduct an additional ablation study utilizing representations from various backbones. In our main experiments, the representation derived solely from AGCN~\cite{Shi_2019_CVPR_twostream} was considered. In this expanded scope, we have incorporated two more GCN-based methods as feature extractors, namely ST-GCN~\cite{yan2018spatial} and CTR-GCN~\cite{chen2021channel}.
}

{The experimental results can be found in Table~\ref{tab: backbones}. 
Our proposed method consistently attains state-of-the-art performance across all three backbones, underscoring the efficacy of our multi-scale and cross-scale matching strategies.}

\begin{table}[t]
\begin{center}
\caption{Experiments with different feature extractors. The experiments were conducted on NTU RGB+D 120
dataset under the Evaluation Protocol 2.
}
\label{tab: backbones}
\resizebox{0.99\linewidth}{!}{
\begin{tabular}{l|c|c|c} 
  \hline 
    Backbone & ST-GCN~\cite{yan2018spatial}  &  CTR-GCN~\cite{chen2021channel}   &  AGCN~\cite{Shi_2019_CVPR_twostream}  \\
  \hline
  ProtoNet~\cite{snell2017prototypical}                  & 59.2  & 61.5   & 60.4  \\
    FEAT~\cite{ye2020fewshot}                   & 59.8   & 62.1   & 61.5  \\
  \hline
  S-Scale (Ours)     & 61.0  & 63.8 & 63.2  \\
  M-Scale (Ours)     &  65.1  & 68.8 & 67.6  \\
  M\&C-Scale (Ours)  & \textbf{66.2}  & \textbf{70.3} & \textbf{68.7}  \\

     \hline

\end{tabular}
}
\end{center}
\end{table}

\noindent\textbf{Visualization Results:}
In order to further show the effectiveness of our proposed multi-scale and cross-scale matching strategies, we provide some visualization examples, as shown in Fig. \ref{fig:visualization}.
For Fig.~\ref{fig:visualization} (a), 
two skeleton samples of the same action "wipe face" were performed by different body parts (two hands vs one hand only)
Similarity-based method (proto) fails in recognition, while our matching-based methods (\textit{e.g.}, `\textbf{S-scale}', `\textbf{M-scale}', `\textbf{M\&C-scale}') can recognize the correct category, showing the effectiveness of the matching-based method.
As shown in Fig.~\ref{fig:visualization} (b), two `use a fan' action samples were performed in different ways. 
The S-scale model fails in recognition since the joint-level representations are different.
However, if we focus on the limb level, these two samples all can be seen as `frequent shaking of arms toward the torso'. 
Therefore, our multi-scale matching-based methods (\textit{e.g.}, `\textbf{M-scale}', `\textbf{M\&C-scale}') successfully recognize the action category, demonstrating the advantage of our proposed multi-scale matching manner.
Fig.~\ref{fig:visualization} (c) shows two skeleton action samples (belonging to `throw') were performed at different motion magnitudes,
and Fig.~\ref{fig:visualization} (d) shows two samples (belonging to `falling ') were performed at different speeds.
Our `\textbf{M\&C-scale}' method can still succeed in recognition in the above challenging two situations, showing that our designed cross-scale matching strategy is able to handle the challenging scenarios where the samples of the same action category can be performed at different magnitudes and different motion paces.

\section{Implementation Details of Compared Few-Shot Learning Methods}
\label{Implementation}

The implementation for the pre-training stage is the same as our single-scale setting in the main manuscript. 
In this section, we focus on the setting for the meta-training stage. 
To achieve the \textit{\textbf{best results}} for those few-shot learning methods \cite{simon2020adaptive,snell2017prototypical,xu2021learning,ye2020fewshot}, we use different settings. 
There are also slight differences in the settings of the same method on different datasets.

\noindent {\textbf{ProtoNet \cite{snell2017prototypical}.}
We utilize the AGCN (a single-scale model, detailed in Figure 5 of the Appendix) as the backbone for ProtoNet. 
During the meta-learning stage, features for ProtoNet learning are extracted using the pre-trained models, specifically from the global pooled feature of the 9th AGCN block, sized at $1 \times 1 \times C^{(9)}$.
Regarding training specifics, for both NTU RGB+D and NTU RGB+D 120 datasets, we set the learning rate to 0.001.
For the PKU-MMD dataset, the initial learning rate is set at 0.0005.
Across all three datasets, we reduce the learning rate by half every 10 epochs. 
For every 5-way 1-shot task, there's a single prototype for each class, meaning each meta task encompasses 5 prototypes.
}

\noindent {\textbf{FEAT \cite{ye2020fewshot}.}
We utilize the AGCN (detailed in Figure 5 of the Appendix) as the backbone for FEAT. During the meta-learning stage, features for FEAT learning are extracted using the pre-trained models, specifically from the global pooled feature of the 9th AGCN block, sized at $1 \times 1 \times C^{(9)}$. 
Diving into the training specifics, across the NTU RGB+D, NTU RGB+D 120, and PKU-MMD datasets, our initial learning rate is set at 0.0005, which we reduce by half every 10 epochs. 
Additionally, there is a weight value to balance the contrastive term in the learning objective. 
Here we set the balance value to 0.1 for NTU RGB+D and NTU RGB+D 120 datasets and 0.01 for the PKU-MMD dataset.  
}

\noindent {\textbf{Subspace \cite{simon2020adaptive}.}
Similar to ProtoNet and FEAT, we also leverage the single-scale model as the backbone for Subspace and the feature for Subspace model learning is also in the size of $1 \times 1 \times C^{(9)}$.  
For NTU RGB+D and NTU RGB+D 120 datasets, we set the learning rate to 0.005.
For the PKU-MMD dataset, the initial learning rate is set at 0.0005.
We cut the learning rate to half every 5 epochs for all three datasets. 
}

\noindent { \textbf{Dynamic Filter \cite{xu2021learning}.}
Consistent with ProtoNet, FEAT, and Subspace, we employ the single-scale model as the backbone for Dynamic Filter [38]. Notably, while Dynamic Filter [38] operates based on feature maps, the input feature for Dynamic Filter learning is in size of $H^{(9)} \times W^{(9)} \times C^{(9)}$, which represents the feature prior to global pooling. 
For NTU RGB+D, NTU RGB+D 120, and PKU-MMD datasets, we set the learning rate to 0.05 and cut the rate by half every 10 epochs.
There's a weighting factor to balance the few-shot classification and global classification objectives. This balance value is set at 0.2 for NTU RGB+D and NTU RGB+D 120 datasets, and 0.1 for the PKU-MMD dataset.
}

{Experiments for all these methods~ \cite{simon2020adaptive,snell2017prototypical,xu2021learning,ye2020fewshot} are optimized using the SGD optimizer with Nesterov momentum (0.9), and the training last for 100 epochs.
}

\section{Notations and Definitions}
{We have included a notation table that delineates each important mathematical symbol and its corresponding definition, thereby enhancing clarity for the reader. This table is presented as Tab.~\ref{tab: notations}.
}

\begin{table}[t]
\begin{center}
\caption{Notations and Definitions.
}
\label{tab: notations}
\renewcommand\arraystretch{1.5}
\begin{tabular}{l|l} 
  \hline 
  \makecell[c]{Notations}  & \makecell[c]{Definitions}  \\
  \hline
   $S$    & Support set   \\
   $Q$    & Query set\\
   $D_{train}$            & Meta-training set\\
   $D_{test}$             & Meta-testing set\\
   $s(\cdot, \cdot)$      & \makecell[l]{The semantic relevance score between \\ two skeleton features}\\
$\mathcal{X}$          & Suppliers\\
$\mathcal{Y}$          & Demanders\\
$x_i$                  & The $i_{th}$ supplier\\
$y_j$                  & The $j_{th}$ demander\\
$OT(\cdot, \cdot)$     & \makecell[l]{The optimal transportation cost between \\ two sets of representations}\\
$\pi$                  & \makecell[l]{The optimal matching flow between \\ two distributions}\\
$r_i$                  & The weight for $i_{th}$ node in suppliers\\
$c_j$                  & The weight for $j_{th}$ node in demanders\\
$d_{ij}$               & \makecell[l]{The pair-wise distance between $i_{th}$ supplier \\ and $j_{th}$ demander}\\
$D_{emd}(\cdot, \cdot)$ & \makecell[l]{The Earth Mover’s Distance between two \\ feature maps}\\
$N $    & The number of skeleton joints   \\
$T $    & The number of frames   \\

  \hline

\end{tabular}
\end{center}
\end{table}

\section{Spatial Pooling}
\label{Pooling}
\begin{figure*}[t]
\begin{center}
\includegraphics[trim=0.2cm 0.3cm 0.2cm 0.1cm,clip,width=0.8\textwidth]{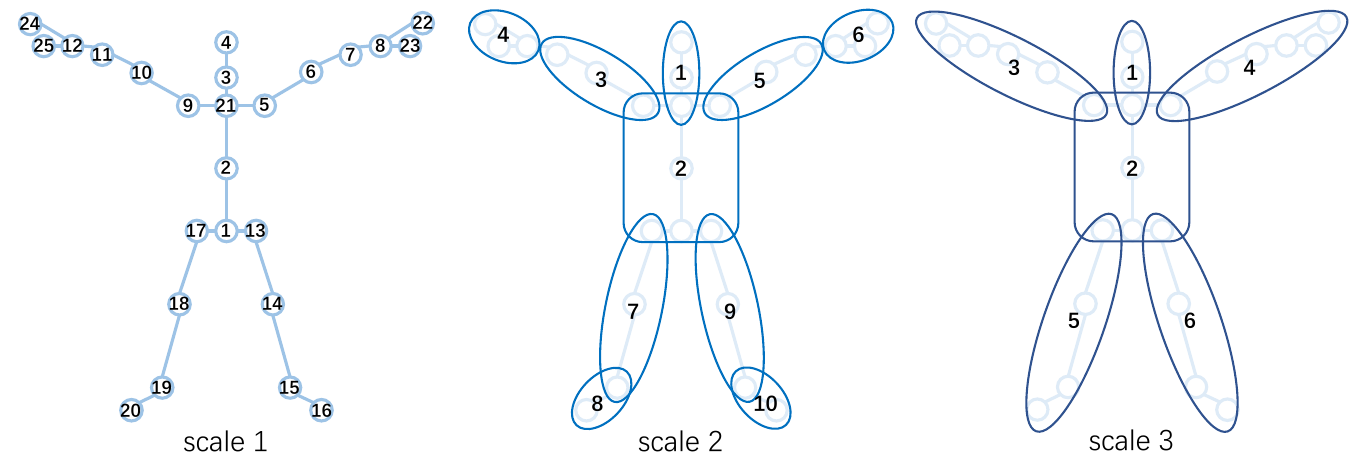}
\end{center}
  \caption{
  Three spatial scales on NTU RGB+D, NTU RGB+D 120, and PKU-MMD datasets. In spatial scale 1, we consider 25 skeleton joints. In spatial scale 2 and spatial scale 3, we consider 10 parts and 6 super-parts, respectively.
  }
\label{fig:spatial pooling}
\end{figure*}

We adopt 3 spatial scales in our work: the joint-level scale, the part-level scale, and the limb(super-part)-level scale, as shown in Fig. \ref{fig:spatial pooling}.
As all three datasets (NTU RGB+D, NTU RGB+D 120, and PKU-MMD) collected skeleton data, which consists of 3D locations of 25 body joints, we consider those 25 skeleton joints for spatial scale 1.
Additionally, in spatial scale 2 and spatial scale 3, we consider 10 parts and 6 super-parts, respectively.
Pooling details from spatial scale 1 to spatial scale 2 and from spatial scale 2 to spatial scale 3 can be found in Tabs. \ref{tab:scale 2} and \ref{tab:scale 3}, respectively.

\begin{table*}[htb]
\parbox{.39\linewidth}{
  \centering
  \caption{
    The pooling details from spatial scale 1 to spatial scale 2.
  }
   \resizebox{.38\textwidth}{!}{
  \begin{tabular}{l|c|c}
    \hline
    Part No. & Name & Joint No. \\
    \hline
    1   & Neck & 3, 4, 21 \\
    2   & Trunk & 1, 2, 5, 9, 13, 17 \\
    3   & Right arm & 9, 10, 11 \\
    4   & Right hand & 12, 24, 25 \\
    5   & Left arm & 5, 6, 7 \\
    6   & Left hand & 8, 22, 23 \\
    7   & Right leg & 17, 18, 19 \\
    8   & Right foot & 19, 20 \\
    9   & Left leg & 13, 14, 15 \\
    10  & Left foot & 15, 16 \\
    \hline
  \end{tabular}
  \label{tab:scale 2}
}
}
\hfill
\parbox{.59\linewidth}{
  \centering
  \caption{The pooling details from spatial scale 2 to spatial scale 3.
  }
 \resizebox{.58\textwidth}{!}{
  \begin{tabular}{l|c|c|c}
    \hline
    Super-Part No. & Name & Part No.& Joint No. \\
    \hline
    1   & Neck & 1  & 3, 4, 21 \\
    2   & Trunk & 2 & 1, 2, 5, 9, 13, 17 \\
    3   & Right upper limb & 3, 4 & 9, 10, 11, 12, 24, 25 \\
    4   & Left upper limb & 5, 6 & 5, 6, 7, 8, 22, 23\\
    5   & Right lower limb & 7, 8 & 17, 18, 19, 20\\
    6   & Left lower limb & 9, 10 & 13, 14, 15, 16 \\
    \hline
  \end{tabular}
  \label{tab:scale 3}
}
}
\end{table*}

\section{Dataset Splitting}
\label{Splitting}

\textbf{NTU RGB+D 120 \cite{ntu120}.}
We follow the official one-shot setting as described in the NTU RGB+D 120 paper \cite{ntu120}. The action classes of the two sets are distinct, which include 100 classes for training and 20 for testing.
The testing set consists of 20 novel classes (i.e. \textit{A1, A7, A13, A19, A25, A31, A37, A43, A49, A55, A61, A67, A73, A79, A85, A91, A97, A103, A109, A115}), and one sample from each novel class is picked as the exemplar.

The following 20 categories are selected: A1 (drink water), A7 (throw), A13 (tear up paper), A19 (take off glasses), A25 (reach into pocket), A31 (pointing to something with finger), A37 (wipe face), A43 (falling), A49 (use a fan (with hand or paper)/feeling warm), A55 (hugging other person), A61 (put on headphone), A67 (hush (quite)), A73 (staple book), A79 (sniff (smell)), A85 (apply cream on face), A91 (open a box), A97 (arm circles), A103 (yawn), A109 (grab other person’s stuff), A115 (take a photo of other person).

As suggested by the original dataset paper \cite{ntu120}, the following 20 samples are selected as the exemplars:
`S001C003P008R001A001', `S001C003P008R001A007', `S001C003P008R001A013', `S001C003P008R001A019', `S001C003P008R001A025', `S001C003P008R001A031', `S001C003P008R001A037', `S001C003P008R001A043', `S001C003P008R001A049', `S001C003P008R001A055', `S018C003P008R001A061', `S018C003P008R001A067', `S018C003P008R001A073', `S018C003P008R001A079', `S018C003P008R001A085', `S018C003P008R001A091', `S018C003P008R001A097', `S018C003P008R001A103', `S018C003P008R001A109', `S018C003P008R001A115'.

\noindent\textbf{NTU RGB+D \cite{ntu60}.} We select 10 novel classes and 10 exemplars from the NTU RGB+D 120 one-shot setting, of which the action label's no. is smaller than 60, as the novel classes and exemplars for the NTU RGB+D dataset.

The following 10 categories are selected: A1 (drink water), A7 (throw), A13 (tear up paper), A19 (take off glasses), A25 (reach into pocket), A31 (pointing to something with finger), A37 (wipe face), A43 (falling), A49 (use a fan (with hand or paper)/feeling warm), A55 (hugging other person). 

The following 10 samples are selected as the exemplars: `S001C003P008R001A001', `S001C003P008R001A007', `S001C003P008R001A013', `S001C003P008R001A019', `S001C003P008R001A025', `S001C003P008R001A031', `S001C003P008R001A037', `S001C003P008R001A043', `S001C003P008R001A049', `S001C003P008R001A055'.

\noindent\textbf{PKU-MMD \cite{pkummd}.}
Similarly, we split PKU-MMD dataset into two parts: the training set (41 classes) and the testing set (10 classes). 
The testing set consists of 10 novel classes, and one sample from each novel class is picked as the exemplar.

The following 10 categories are the novel classes: A1 (bow), A6 (clapping), A11 (falling), A16 (hugging other person), A21 (pat on back of other person), A26 (punching/slapping other person), A31 (rub two hands together), A36 (take off glasses), A41 (throw), A46 (typing on a keyboard).

The following 10 samples are the exemplars:
`0003-L\_A\_1', `0003-L\_A\_6', `0002-L\_A\_11', `0005-L\_A\_16', `0005-L\_A\_21', `0005-L\_A26', `0002-L\_A\_31', `0003-L\_A\_36' , `0002-L\_A\_41', `0003-L\_A\_46'.   

The videos in the PKU-MMD dataset are untrimmed, so we need to trim videos to the one-action segment level based on the given starting time and ending time.
While the videos' filenames contain only the part before the first `\_' of exemplars' filenames, take the `0003-L\_A\_1' as an example, the original filename is `0003-L'.
Since we trim the video, we add the action category number in the filename, here `A\_1' in `0003-L\_A\_1' means the corresponding segment of action category \textbf{1} in the video `0003-L'.

\ifCLASSOPTIONcaptionsoff
  \newpage
\fi

\end{document}